\documentclass[sn-mathphys-num]{sn-jnl}

\usepackage{graphicx}%
\usepackage{multirow}%
\usepackage{amsmath,amssymb,amsfonts}%
\usepackage{amsthm}%
\usepackage{mathrsfs}%
\usepackage[title]{appendix}%
\usepackage[table]{xcolor}%
\usepackage{textcomp}%
\usepackage{manyfoot}%
\usepackage{booktabs}%
\usepackage{algorithm}%
\usepackage{algorithmicx}%
\usepackage{algpseudocode}%
\usepackage{listings}%
\usepackage{multirow}%
\usepackage{nicematrix}%
\usepackage{adjustbox}%
\usepackage{booktabs}

\theoremstyle{thmstyleone}%
\theoremstyle{thmstyletwo}%

\theoremstyle{thmstylethree}%

\begin{document}


\title[Article Title]{Robust and Explainable Framework to Address Data Scarcity in Diagnostic Imaging}


\author[1]{\fnm{Zehui} \sur{ Zhao}}\email{zehui.zhao@hdr.qut.edu.au}

\author*[2]{\fnm{Laith} \sur{Alzubaidi}}\email{l.alzubaidi@qut.edu.au}

\author[1]{\fnm{Jinglan} \sur{Zhang}}\email{jinglan.zhang@qut.edu.au}

\author[3]{\fnm{Ye} \sur{Duan}}\email{duan@clemson.edu}

\author[4]{\fnm{Usman} \sur{Naseem}}\email{usman.naseem@mq.edu.au}

\author[2]{\fnm{Yuantong} \sur{Gu}}\email{yuantong.gu@qut.edu.au}

\affil[1]{\orgdiv{School of Computer Science}, \orgname{Queensland University of Technology}, \orgaddress{\street{2 George Street}, \city{Brisbane}, \postcode{4000}, \state{Queensland}, \country{Australia}}}

\affil*[2]{\orgdiv{School of Mechanical, Medical, and Process Engineering}, \orgname{Queensland University of Technology}, \orgaddress{\street{2 George Street}, \city{Brisbane}, \postcode{4000}, \state{Queensland}, \country{Australia}}}

\affil[3]{\orgdiv{School of Computing}, \orgname{Clemson University}, \orgaddress{ \city{Clemson}, \postcode{29631}, \state{South Carolina}, \country{USA}}}

\affil[4]{\orgdiv{School of Computing}, \orgname{Macquarie University}, \orgaddress{ \city{Sydney}, \postcode{2109}, \state{New South Wales}, \country{Australia}}}


\abstract{Deep learning has significantly advanced automatic medical diagnostics and released the occupation of human resources to reduce clinical pressure, yet the persistent challenge of data scarcity in this area hampers its further improvements and applications. To address this gap, we introduce a novel ensemble framework called `Efficient Transfer and Self-supervised Learning based Ensemble Framework' (ETSEF). ETSEF leverages features from multiple pre-trained deep learning models to efficiently learn powerful representations from a limited number of data samples. To the best of our knowledge, ETSEF is the first strategy that combines two pre-training methodologies (Transfer Learning and Self-supervised Learning) with ensemble learning approaches. Various data enhancement techniques, including data augmentation, feature fusion, feature selection, and decision fusion, have also been deployed to maximise the efficiency and robustness of the ETSEF model. Five independent medical imaging tasks, including endoscopy, breast cancer, monkeypox, brain tumour, and glaucoma detection, were tested to demonstrate ETSEF's effectiveness and robustness. Facing limited sample numbers and challenging medical tasks, ETSEF has proved its effectiveness by improving diagnostics accuracies from 10\% to 13.3\% when compared to strong ensemble baseline models and up to 14.4\% improvements compared with published state-of-the-art methods. Moreover, we emphasise the robustness and trustworthiness of the ETSEF method through various vision-explainable artificial intelligence techniques, including Grad-CAM, SHAP, and t-SNE. Compared to those large-scale deep learning models, ETSEF can be deployed flexibly and maintain superior performance for challenging medical imaging tasks, showing the potential to be applied to more areas that lack training data.}




\maketitle

\section{Introduction}
Deep Learning (DL), a cutting-edge branch of Machine Learning (ML) technology, has demonstrated remarkable success in various medical imaging modalities, including radiology images \cite{fernandez2023deep, saba2019present, ardila2019end}, dermatology and ophthalmology images \cite{esteva2017dermatologist,mazhar2023role, panahi2024simplified}, and 3D images derived from multiple modalities \cite{hatamizadeh2022unetr, khader2023denoising}. These state-of-the-art DL models have showcased their ability to perform tasks that expert clinicians traditionally perform, including disease classification, segmentation, localisation, and diagnosis. The integration of DL into medical systems holds the promise of enhancing diagnosing efficiency and releasing the pressure of healthcare services, thus improving accessibility and coverage in public healthcare \cite{chan2020deep}.

However, despite the significant strides made in DL, the persistent challenge of data scarcity impedes its widespread application in challenging medical settings. DL algorithms rely on complex multi-layer architectures with millions of parameters to capture intricate data distributions, which are difficult to acquire in many medical domains \cite{alzubaidi2023survey}. The scarcity of training data in this field can be attributed to several factors: (i). The high cost associated with medical imaging tests and the expensive process of annotating medical data through specialised expertise \cite{tajbakhsh2020embracing}. (ii). concerns regarding patient privacy, which may hinder the sharing and reuse of medical datasets \cite{kaissis2021end, dhar2023challenges}. (iii). The diverse modalities and high-resolution nature of medical images demand meticulous data collection and reduce data generalisation \cite{zhou2021review}. (iv). The infrequent occurrence of rare diseases limits the availability of the corresponding data samples \cite{banerjee2023machine}. Consequently, these factors lead to data scarcity issues and pose a significant barrier to the further advancement and application of DL techniques in the medical field, increasing the risk of overfitting and unexpected performance degradation during model deployment \cite{bansal2022systematic}. 

In response to the challenge of data scarcity, one simple approach is to invest more resources in obtaining expert annotations and clinical samples while developing tailored models from scratch to better align with the target data distribution \cite{javaid2022significance, chen2022recent}. However, these solutions entail significant financial and temporal burdens, potentially limiting the practicality and lifespan of DL models in real-world scenarios \cite{petersen2022responsible}. Alternatively, researchers have explored practical solutions to enhance model performance, which can be broadly categorised as data-level and model-level approaches. For instance, data augmentation serves as a popular data-level solution and addresses data scarcity by generating synthetic data to enlarge the training data size \cite{shorten2019survey, kebaili2023deep}, while pre-trained learning methods represent another popular direction that aims at enhancing the model's knowledge base and feature extraction capabilities \cite{laurer2024less}.

In this study, we focus on pre-trained learning and ensemble learning methods. Transfer Learning (TL) stands out as a widely adopted approach among all the pre-trained learning methodologies. The normal TL model aims to improve the learning of target domain data by transferring knowledge learnt from different but related source domains \cite{zhuang2020comprehensive}. This method allows users to reuse the available trained models and existing datasets to enhance the model training process instead of spending time and money to design a new model from scratch \cite{iman2023review}. So far, much research has proved the TL method's value in improving the target model's training efficiency, performance, generalisation ability, and data reuse rate \cite{malik2020comparison, kathamuthu2023deep, salehi2023study}. Despite its successes, the TL knowledge transfer process faces a significant risk called negative transfer when transferring knowledge between dissimilar domains \cite{zhang2022survey}. The dissimilarity between domains can be represented as a difference in feature distribution or representative symptoms, and this difference may cause heavy degradation in the model's performance and robustness. Furthermore, existing TL methods are mainly based on supervised learning and rely heavily on human instructions, which require additional support from massive source data and may limit their applicability and scalability in areas lacking source domain data \cite{niu2020decade}.

In contrast, self-supervised learning (SSL) has recently emerged as a promising unsupervised pre-training approach and has attracted the attention of many researchers. Unlike TL methods, SSL methods utilise unlabelled data to extract rich features through well-designed pretext task training, thus reducing data preparation and annotation efforts \cite{krishnan2022self}. Furthermore, by using augmented training data as pseudo-labels, SSL facilitates simulated supervised training to help the model achieve better performance behaviour \cite{navarro2021evaluating}. In recent years, the SSL method has achieved a series of successes in the medical field and can even compete with or overpass the TL-based algorithms in some scenarios \cite{zhao2023comparison}. However, although SSL has shown promising potential in the medical domain, it is also beset by challenges, including negative transfer risk such as TL methods, high computational consumption compared to other supervised learning paradigms \cite{ericsson2022self2}, the unreliability of lacking strong supervision signals, and the weak ability to capture changeable characteristic information through pretext task training \cite{ericsson2022self1}.

To address the inefficiencies and risks associated with existing pre-trained learning methods, researchers attempted to develop multiple publicly available large-scale datasets as source domains to support pre-trained learning across various fields. A typical sample is ImageNet \cite{deng2009imagenet}, a publicly available dataset containing 14 million well-annotated image data collected from general areas, including animals, humans, cars, plants, etc. Despite the success of ImageNet-based pre-trained models in many application areas, there have been concerns regarding the suitability of ImageNet and other generic source datasets in medical applications due to the differences in feature distributions between general images and professional medical images. As a response, Tan and Niu's team \cite{tan2017distant, niu2021distant} has proposed a multi-level pre-training strategy by pre-training the TL model in generic datasets first and retraining it with an intermediate source domain that is close to the target domain, bridging the distribution gap between the large-scale generic source domain and professional target domain to avoid negative transfer. Furthermore, Azizi et al. \cite{azizi2021big, azizi2023robust} developed a unified pre-training strategy to combine TL-based weights and SSL pre-training paradigms to enhance model capability against unseen data. Taking advantage of a large-scale non-medical knowledge base and retraining the model on an intermediate medical source domain, their proposed method has shown superior performance and generalisation against unseen medical data. However, their work lacks adequate visual analysis and explainable evidence to support the decision-making process, risking biased predictions based on unrelated representations. To address this issue, we employed ensemble learning techniques combined with DL and pre-trained algorithms, as previous research shows it can effectively enhance model performance and robustness \cite{yang2017ensemble,zhou2021domain, mungoli2023adaptive}. Utilising multiple base models trained on the same tasks and complementary information from each model, the ensemble learning strategy is able to expand the weak models' generalisation ability and strengthen them into a more powerful integrated model. 

\begin{figure}[htbp]
\centerline{\includegraphics[width=0.98\textwidth]{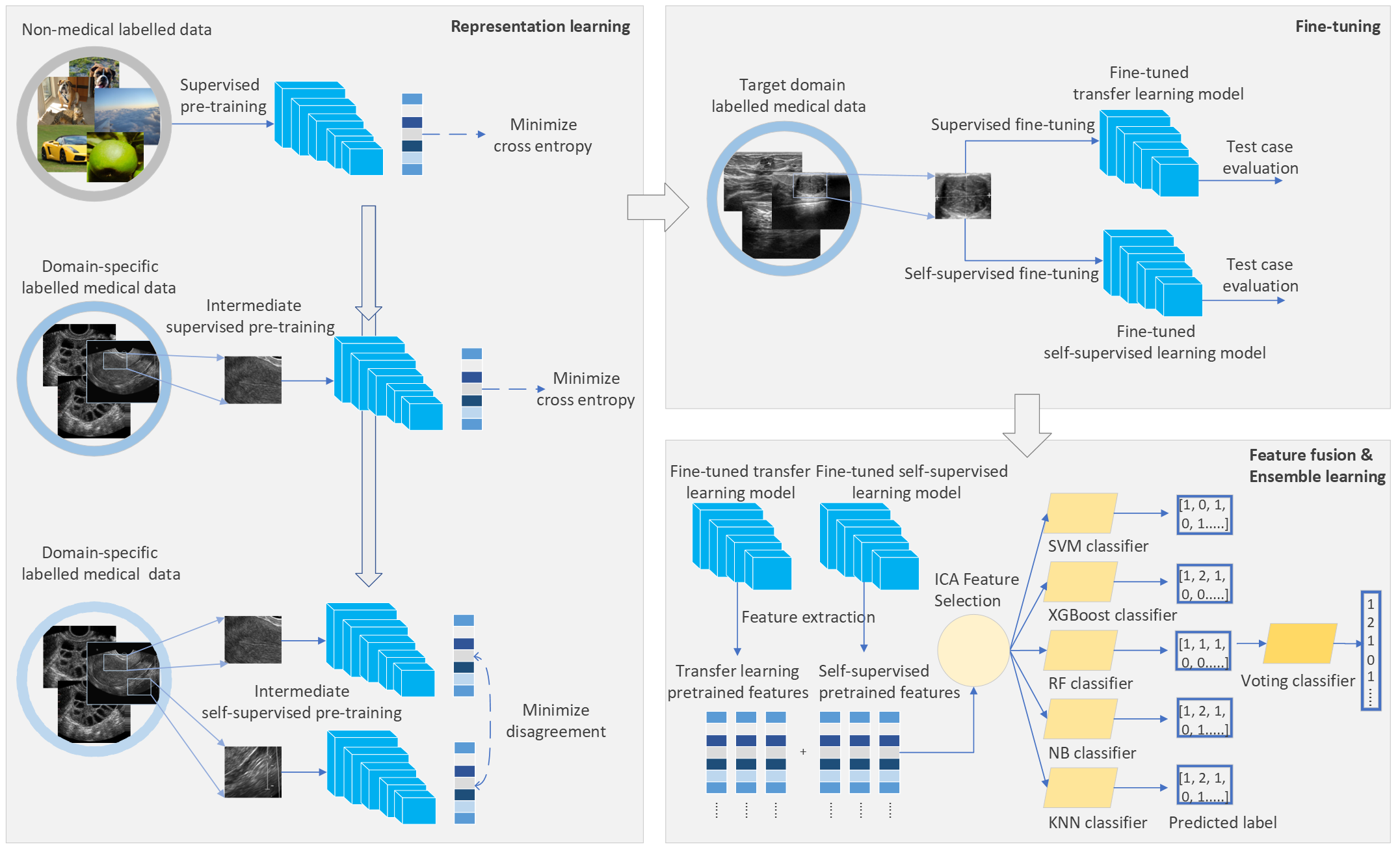}}
\caption{\textbf{Overview of the ETSEF approach workflow} ETSEF initialises the base model by leveraging pre-trained knowledge from ImageNet using supervised learning. Next, two separate pre-trained models utilise TL and SSL methods to learn representations from the same intermediate data. These models are then fine-tuned for specific downstream medical imaging tasks. Subsequently, features are extracted from the models for concatenation and normalization. We use five ML classifiers to learn the feature distribution and make predictions. Finally, a voting classifier aggregates predictions and makes a final decision based on the most consistent outcomes.}
\label{Overview of framework}
\end{figure}

Inspired by the pressing need for a robust and data-efficient approach to medical DL, we build upon recent advancements in this field to introduce the ETSEF, a framework that proposes an ensemble strategy based on multiple pre-training methods to craft powerful models for medical imaging tasks (see Fig. \ref{Overview of framework} for an overview of ETSEF). Through extensive experimentation, we demonstrated the effectiveness and robustness of ETSEF across various clinical scenarios. The primary objective of this framework is to learn reusable and transferable representations and enhance task learning in data-scarced clinical environments. Our approach leverages intermediate medical source domains to support downstream task learning and constructs robust base feature extractors using limited target samples. Subsequently, the ensemble strategy selects and merges task-relevant features from each base model and employs higher-level ML classifiers such as Random Forest, eXtreme Gradient Boosting, and Naive Bayes as feature classifiers to refine the understanding of the target feature distribution and facilitate informed predictions. Finally, a voting classifier aggregates the weighted predictions of the preceding feature classifiers to arrive at a consensual decision. 

In contrast to previous pre-training methods that focus on individual medical imaging domains such as chest-X-ray \cite{reddy2023multi, wani2023osteoporosis}, brain Computed Tomography (CT) scan \cite{tang2022self, li2021transfer, wolf2023self}, Magnetic Resonance Imaging (MRI) interpretation \cite{ghassemi2020deep} etc, our method emphasises generalisation ability across various medical imaging domains. To our knowledge, ETSEF stands as the first instance of an ensemble learning framework that combines two independent pre-training methodologies. This approach harnesses multiple DL architectures and pre-training techniques to learn comprehensive symptom representations from the target sample, bridging the gap between disparate domains and minimising the need for customised model designs on different data domains. Notably, our strategy follows Azzi and Niu's work \cite{niu2021distant,azizi2023robust}, leveraging generic source pre-trained weights as a knowledge base and further pre-trained the base models with intermediate medical source data to enhance the base model's performance and avoid potential negative transfer risks. Furthermore, our experiment also encompasses various ensemble learning techniques to offer more insight and recommendations for developing a powerful and efficient ensemble model. 

Additionally, we prioritise the explainability of our framework, recognising the critical role of model security in medical artificial intelligence model learning. To enhance transparency, we employed Gradient-weighted Class Activation Mapping (Grad-CAM) \cite{selvaraju2017grad} and SHapley Additive exPlanation (SHAP) \cite{ribeiro2016should} techniques to our pre-trained base models to provide visual explanations and supportive evaluation. With the support of these explainable artificial intelligence techniques, we highlight the model's attention and focus areas, offering visual insights into the model's decision-making process. Furthermore, we utilised t-SNE analysis \cite{van2008visualizing} to assess the feature classification performance. With the help of the visualised feature maps, we are able to evaluate the extracted feature quality and gain more understanding of final predictions.

In this section, we introduce our motivation for conducting this study and provide an overview of the ETSEF method. The second section presents our strategy design, clinical environments, performance comparison with baselines and state-of-the-art models, and visual analysis. The third section discusses our method design, while the fourth section details the implementation and literature background of the ETSEF. The final section concludes our findings and outlines future directions.

\section{Results}\label{sec2}

In addition to outlining the overall workflow, this section offers valuable insights into the ETSEF method, including the methodology and component design, the clinical data environment setup, the model performance analysis, and the model predictions' visual analysis. Both ensemble learning and pre-trained models are essential components of our framework, and our detailed exploration of these topics serves to illuminate the framework's design while providing a comprehensive guide for leveraging our findings. 

\subsection{An Ensemble Framework that Combines Two Pre-trained Learning Methods}

The primary objective of our method is to develop a model that achieves high performance across various medical imaging tasks with limited training data. To reduce the negative influence of data scarcity, we utilised multiple pre-trained models as base models and combined them into a robust ensemble model. These base models are derived from two popular pre-training methods: transfer learning and self-supervised learning. In addition, we incorporated feature fusion, feature selection, and weighted voting techniques in the ensemble learning process to ensure the performance and stability of the final model. 

\textbf{Supervised transfer learning.} Pre-trained supervised TL models have demonstrated their effectiveness with various medical imaging tasks \cite{asif2023enhanced, chow2023quantitative}, here we follow the theoretical groundwork established by Azizi et al. \cite{azizi2023robust} to leverage the publicly available pre-trained representations as knowledge foundation and subsequently use the double fine-tuning technique to train our predictors to map related labelled medical data before finally fine-tuning on the target domain. Notably, one major advantage of double fine-tuning is the flexibility in selecting intermediate source data, which need not strictly adhere to the same disease types or large dataset sizes. Additionally, in scenarios where labelled source data is unavailable, our method can provide an alternative option to utilise related unlabeled medical data as the source domain and pre-train solely with the SSL algorithm. The experiment results demonstrate the efficacy of using source domain data from similar medical imaging sources, even with a relatively small dataset size that contains only hundreds of samples. Compared to using large-scale datasets for pre-training, our approach proves effective in improving target task performance without adding much computational cost.

In the representation learning stage, we initialise predictors $h(\cdot)$ by employing multiple pre-trained DL models with ImageNet weight. With most model layers unfrozen, we retrain them for related medical representations by minimising cross-entropy loss on the intermediate source domain data. Specifically, we selected Xception \cite{chollet2017xception}, InceptionResNetV2 \cite{szegedy2017inception}, and MobileNetV3 \cite{howard2019searching} as backbone encoder networks for their ease of deployment, superior performance, and suitable depth for limited medical data environments. In the fine-tuning stage, each backbone network $f(\cdot)$ is augmented with a modified classification head $g(\cdot)$ to create the final predictor $h = g \circ f$ and learn mapping representations to the target-specific label space. The trained medical knowledge from the previous stage is partially frozen during this step for the convenience of knowledge reuse. Given labelled training examples $(x_1, y_1),...,(x_n,y_n)$ from the downstream medical dataset, the predictor $h(\cdot)$ is fine-tuned to minimise cross-entropy and learn domain-specific representations to support subsequent ensemble training. This process is repeated until each backbone network has acquired the necessary knowledge for the target samples.

\textbf{Contrastive Self-supervised Learning.} Model training using the SSL method allows the base model to effectively learn visual representations from unlabeled medical images, providing a unique perspective distinct from supervised transfer learning. To harness the advantages of SSL and complement the supervised TL method, we adopted SimCLR \cite{chen2020simple, chen2020big}, an SSL algorithm based on contrastive learning to train another group of base models. SimCLR learns representations by maximising agreement between pseudo-labels augmented from the same training samples with different views and original training examples. 

In the representation learning stage, the initial predictors $f_\theta(\cdot)$ employ DL architectures similar to the pre-trained backbone encoder networks used in the TL setting, and a two-layer projection head is used to project the representation to a 128-dimensional latent space. However, to maximise the diversity of base models, we select a disparate architecture ResNet101 \cite{he2016deep}, along with MobileNetV3 and EfficientNetB0 \cite{tan2019efficientnet} as backbone encoders. Given the majority of predictor layers keep frozen, the SimCLR algorithm learns representations $z_{2k-1}$ and $z_{2k}$ from augmented views of the training sample and computes contrastive loss objectives. With a mini-batch of encoded examples, the contrastive loss between a positive pair of examples i, j (different augmentations of the same image) is given as follows:    \begin{small}
$$ l_{i,j} = -\log\frac{exp(sim(z_i,z_j)/\tau)}{\sum^{2N}_{k=1}l_{[k\neq i]}exp(sim(z_i,z_k)/\tau) } $$
\end{small} where cosine similarity computes distances between two vectors and $\tau$ is a scalar that denotes the temperature. 

In the fine-tuning stage, the encoder network $f_\theta(\cdot)$ is replaced with a new classification head $g_\theta(\cdot)$ to map the predictions to the target class labels and compose the final predictor $h_\theta = g_\theta \circ f_\theta$. This predictor is trained to minimise cross-entropy loss using the same labelled training examples $ (x_1, y_1),...,(x_n,y_n) $ providing target domain representations for subsequent ensemble training. Both TL and SSL training processes are influenced by multiple hyperparameters, including optimiser type, learning rate, training epochs, batch size, and weight decay, and we provided more details in the Methods section. 

\textbf{Feature Fusion methods.} In the feature fusion stage, we combine the extracted features from pre-trained TL and SSL base models using feature fusion techniques. Feature fusion is a process of fusing features from different layers or branches of different sources or models to create a more informative feature representation \cite{mangai2010survey}. Facing the challenging medical imaging tasks, although recent state-of-the-art pre-trained learning models using TL or SSL methods are able to capture useful features from the target data samples, it is hard for them to handle the wide variability and scalability of the data in a domain where extremely limited data are available. Thus, the feature fusion method plays a crucial role in our framework to expand the diversity and complementarity of the individual models, and promote the robustness and accuracy of the overall decision-making process \cite{zhang2019feature}. Under this situation, three main feature fusion methods are widely applied: Concatenation, Element-wise addition and multiplication, and Linear and nonlinear transformation.

\begin{itemize}
\item Concatenation simply combines the feature vectors of multiple sources by appending them together. This method maximises the retention of original information but may cause high-dimensional feature representations and raise feature complexity.
\item Element-wise addition and multiplication method first computes the element-wise sum or product of the feature vectors separately before combining them to reduce the dimensionality of the fused feature. However, this method may cause a loss of original information when calculating and selecting them.
\item Linear and nonlinear transformation method works by introducing additional linear or nonlinear functions to project the feature vectors into a common feature space to select the most relevant features. Examples of these methods include Principal Component Analysis (PCA), Independent Principal Component Analysis (ICA), and Linear Discriminant Analysis (LDA).
\end{itemize}

Using feature fusion, we aim to preserve the most valuable features for the target domain tasks and optimise fusion performance across the base pre-trained models. Consequently, we explore the effectiveness of concatenation, PCA, ICA, and LDA methods in enhancing feature quality. Among these methods, concatenation served as the baseline method since it retains all the information about the characteristics and allows comparison with other normalisation methods. PCA, an unsupervised feature normalisation technique, captures maximum data variance and maintains linearly uncorrelated variables known as principal components \cite{jolliffe2016principal}. In contrast, LDA is a supervised feature normalisation method that is similar to PCA, LDA incorporates class labels and aims to maximise class separation while minimising class variance \cite{choubey2020comparative}. ICA is also an unsupervised feature normalisation technique that aims to separate a multivariate signal into additive independent components \cite{calhoun2009review}. It assumes that the observed data are mixtures of independent sources, and it tries to recover the original sources. 

Due to content constraints and research focus, we present experimental observations rather than exhaustive details. Our findings indicate superior performance of unsupervised transformation methods such as PCA and ICA compared to the baseline concatenation method, while supervised LDA performs comparatively worse. We attribute this performance gap in supervised methods to their retention of only a limited number of features matching class label numbers, leading to a significant loss of feature information and poorer performance. Additionally, concatenation is deemed unsuitable as it may increase feature dimensions and complexity due to coincidences in extracted features from multiple base models. Consequently, we utilised the concatenation method to keep all the feature information first and then applied the ICA method for its ability to maximise independent feature representations and preserve the unique attention views from diverse base models. 

\textbf{Unified Ensemble Learning using Pre-trained Deep Learning Architectures.} In the ensemble learning stage, previously pre-trained base models are integrated using ensemble learning technology that combines multiple base models to create a more robust and powerful model \cite{zhou2012ensemble}. This approach aims to leverage the strengths of diverse models while mitigating their weaknesses, resulting in an improved overall model \cite{dong2020survey}. 

In recent years, ensemble-based proposals have maintained steady growth and serve as a state-of-the-art method in many fields \cite{gonzalez2020practical, wu2023application, louk2023dual}. However, traditional ensemble learning methods that use weak base models to form a more powerful model have been found to be computationally inefficient and less reliable. The effectiveness of ensemble learning depends heavily on the performance of its base models. Errors in a single base model can be amplified within the ensemble learning process, leading to instability issues in the overall model \cite{rokach2016decision}. In this circumstance, the combination of powerful DL models with a team technique has shown potential to address this issue and achieved great success in the medical field \cite{ khozeimeh2022rf, bhuiyan2023new, kaya2024feature}. Our method uses double-fine-tuned pre-training models to reinforce the performance of the base models, and we referred to three main ensemble learning techniques that are commonly applied:

\begin{itemize}
\item Bagging \cite{breiman1996bagging} trains multiple base models independently on different subsets of the same target data using random sampling with replacement. The final model's predictions are determined by aggregating the individual base models' predictions through voting. 
\item Boosting \cite{schapire2013boosting} employs a sequence of base models to iteratively correct the errors of its predecessors and update them dynamically. Final predictions are also obtained through a weighted vote of the base models.
\item Stacking \cite{wolpert1992stacked} combines the outputs of multiple base models using a higher-level model, often referred to as a meta-model. The meta-model is trained on a new dataset comprising the base models' predictions, subsequently making new predictions as output.
\end{itemize}

We employed a modified method based on the stacking method to learn representations from pre-trained base models effectively. This involved training fused features at a higher model level with five ML models, including K-Nearest Neighbours (KNN), eXtreme Gradient Boosting (XGBoost) \cite{chen2016xgboost}, Support Vector Machine (SVM), Naive Bayes (NB), and Random Forest (RF). Subsequently, a weighted vote of the ML models' predictions was conducted, with each model contributing equally to the final decision and taking the majority voted predictions. The training process of the ML model is influenced by independent parameters such as max-depth, number of neighbours, random state, etc., which is based on the specific requirement of the ML model. The following method section will discuss these parameters and their impact in further detail.

\subsection{Clinical Evaluation Settings}

In this study, our ETSEF method was mainly trained and evaluated for four domain-specific tasks that belong to independent medical imaging modalities (dermatology photography, gastrointestinal tract, breast ultrasound and chest radiography), spanning from colourful domains to grey-scale domains. Each base model has been trained through an initial source domain (ImageNet), a related medical dataset $D_{in}$ as the intermediate source domain, and a target domain dataset $D_{tar}$. To facilitate the final training and testing, all data sets from the target domain were randomly split into train and test sets with a ratio of 8: 2, and we further enhanced the data of the train sets to reduce the potential influence of data imbalance problems. Moreover, we also included an out-of-distribution task to test the generalisability of the ETSEF strategy. Here, we provide an overview of each task and their corresponding datasets.

\textbf{Task 1: Dermatology MonkeyPox Classification.} Recently, the World Health Organization highlighted the rapid spread of MonkeyPox across Europe and America. During the initial peak outbreak phase of MonkeyPox, a significant challenge emerged in distinguishing MonkeyPox patients from those with other pox or skin diseases. This task aims to train an efficient model to detect MonkeyPox patients quickly. The intermediate source dataset $D_{in}^{T1}$ is a publicly available dataset called the Monkeypox Skin Images Dataset (MSID) \cite{bala2023monkeynet}, which compromised 770 unique cases collected from internet-based sources. $D_{in}^{T1}$ includes skin and body images of patients and healthy individuals of various sexes, ages, and races. This dataset covers three diseases—MonkeyPox, Chickenpox, and Measles, along with images of healthy individuals. The target dataset $D_{tar}^{T1}$ for the task is the MonkeyPox Skin Lesion Dataset Version 2.0 (MSLDv2) \cite{ali2023web}. Specifically, $D_{tar}^{T1}$ focuses on various pox diseases and contains 755 original skin lesion images from 541 distinct patients. $D_{tar}^{T1}$ includes six distinct classes, namely Monkypox (284 images), Chickenpox (75 images), Measles (55 images), Cowpox (66 images), Hand-foot-mouth disease or HFMD (161 images), and Healthy (114 images). To address the data imbalance between the Monkeypox class and the other classes, we applied data augmentation to the five smaller classes, balancing their sample numbers to mitigate the influence of the imbalance. The augmented training set comprises 1,414 samples, with approximately 240 samples per class. Compared to the intermediate source dataset $D_{in}^{T1}$, the target dataset $D_{tar}^{T1}$ share the same disease types such as MonkeyPox, ChickenPox and Measles. Moreover, the target $D_{tar}^{T1}$ is more complex and covers additional cases, including CowPox and Hand, foot, and mouth disease; the increased disease classes make the target classification task more challenging and suitable for testing the performance of the ETSEF in complex scenarios.

\textbf{Task2: Gastrointestinal Endoscopy Classification.} Gastrointestinal (GI) endoscopy is an active field of research driven by the high incidence of lethal GI cancers. Early GI cancer precursors are often missed during endoscopic surveillance, necessitating the integration of experienced endoscopists’ knowledge to improve disease classification accuracy. The objective of this task is to train a robust model to classify anatomical landmarks and clinically significant findings from GI tract images. For this purpose, we utilise an intermediate source dataset $D_{in}^{T2}$ from the Medico Multimedia Task at MediaEval 2018 (Medico 2018) \cite{jha2021comprehensive} to provide a useful knowledge base. $D_{in}^{T2}$ comprises 5,293 GI images, including anatomical landmarks, pathological findings, polyp removal cases, and normal cases, categorised into 16 classes. The target dataset $D_{tar}^{T2}$ we use in this task is Kvasirv2 \cite{pogorelov2017kvasir}, which contains 8,000 GI tract images of anatomical landmarks and pathological findings. $D_{tar}^{T2}$ is a balanced dataset across eight classes, with 1,000 samples per class. Data augmentation was also applied to this dataset to evaluate the effectiveness of this technique within our framework, doubling each class sample while maintaining balance, leaving a train set with 12,800 samples and a test set with 1,600 samples. Notably, the Medico 2018 dataset also includes some samples from the Kvasir version 1.0 dataset, and we compared $D_{in}^{T2}$ and the target test set to ensure there are no data duplication issues.

\textbf{Task3: Breast Ultrasound Classification.} Breast cancer is one of the most common causes of death among women worldwide. Early detection is crucial in reducing mortality rates. This task aims to classify whether a patient has breast cancer and determine if the cancer is benign or malignant. Due to the lack of related source datasets, we selected a publicly available ultrasound dataset from Kaggle, the Polycystic Ovary Syndrome (PCOS) detection dataset \cite{chou2021ultra}, as our source dataset $D_{in}^{T3}$. Specifically, $D_{in}^{T3}$ is a binary dataset with 1,924 images categorised into two classes: infected and not infected by PCOS. We chose this dataset because it uses ultrasound imaging, providing similar grey-scale medical representations. The target dataset $D_{tar}^{T3}$ for this task is the Breast Ultrasound Dataset (BUSI) \cite{al2020dataset}. The BUSI dataset comprises 780 images collected from 600 female patients, with ground truth labels provided by expert annotation. The samples are categorised into three classes: Normal (133 samples), Benign (437 samples), and Malignant (210 samples). To address the class imbalance, we performed data augmentation on the normal and malignant samples, resulting in 279 images for the normal class and 340 images for the malignant class in the augmented training set. Additionally, the original dataset includes 798 masked ground truth images for segmentation training, which we excluded as they are not relevant to our target task.

\textbf{Task4: Brain Tumour Detection} Brain tumours are among the most aggressive diseases, and MRI is the best technique for detecting them. This task aims to train models on two MRI datasets to diagnose different brain tumours and determine if a patient has the disease. The source domain dataset $D_{in}^{T4}$ is a publicly available Kaggle dataset called Br35H \cite{Hamada2020br35h}. Br35H contains 3,060 MRI image samples divided into two classes: with and without brain tumours. Similarly, the target dataset $D_{tar}^{T4}$ for this task is also a tumour image dataset called Brain Tumor Classification (BTC) dataset \cite{sartaj2020brain}, which includes 3,264 image samples categorised into three types of tumours (glioma, meningioma, and pituitary) and non-tumour cases. The sample number of each class is generally balanced among the three tumour classes, but the non-tumour class is underrepresented. To address this issue, we performed data augmentation on the non-tumour samples, resulting in a more balanced sample distribution with approximately 750 images for each class in the training set. The  $D_{in}^{T4}$ and $D_{tar}^{T4}$ share a similar feature distribution and data size, while $D_{tar}^{T4}$ contains more classes and has a more imbalanced class distribution, making it a challenging task to face.

\textbf{Task5: Out-of-distribution Disease Detection.} To test the generalisation ability of ETSEF, we conduct data-efficient performance testing under realistic out-of-distribution (OOD) settings, where OOD refers to newly appeared and previously unseen clinical environments. Leveraging the ensemble learning structure, our ETSEF strategy can utilise any trained base models as feature extractors to process unseen data without requiring additional retraining or fine-tuning steps. The ML classifiers in the ensemble learning stage build correlations between these extracted feature vectors and map them to predicted labels. In addition to model performance, the generalisation ability of the model is particularly valuable in the medical field, where varying imaging techniques often cause significant feature distribution shifts between different datasets. For this purpose, we selected a glaucoma dataset called EyePACS-AIROGS-light-V2 (EyePacs) \cite{kiefer2023catalog} as our target dataset. We utilised pre-trained base models from the previous task $T_2$ as feature extractors for this task, as they were also trained on colourful medical image datasets, and the representation knowledge they learned from the gastrointestinal endoscopy classification task is markedly different from those for glaucoma images. Moreover, no further source domain datasets or fine-tuning steps were used since we want to keep the feature difference between pre-trained weights with the target dataset to test the trained ETSEF model's generalisation ability under extreme environments. The EyePacs dataset can be used to conduct a binary classification task with 6,000 samples. Following the overall protocol, we split the target dataset into training and testing sets, using 1,000 samples in the test set to perform the performance evaluation.

\textbf{Experimental setup.} To test the clinical performance and analyse the effectiveness of ETSEF under different clinical settings, standard supervised baseline models were trained for $T1-4$ without using any pre-trained weights. These baseline models, which utilised the same base DL architectures, were then subjected to feature fusion and ensemble learning techniques to create a scratch baseline ensemble model. Additionally, we built another group of TL-based ensemble models and SSL-based ensemble models using single pre-trained method-based models for comparison. The rationale for building separate pre-trained ensemble models is to establish a baseline with one pre-training-method ensemble model, thereby demonstrating the performance of fusing multiple pre-training methods.

\begin{figure}[htbp]
\centerline{\includegraphics[width=0.98\textwidth]{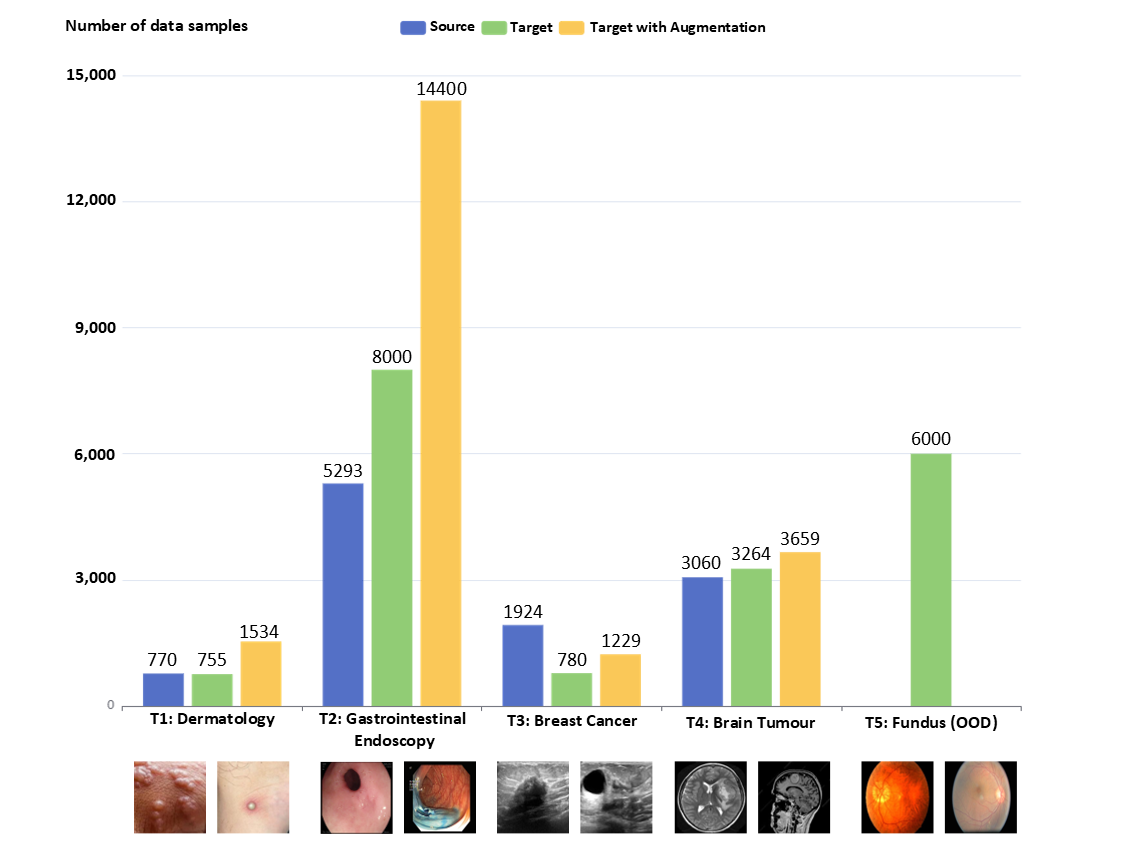}}
\caption{\textbf{Overview of datasets for each task} The trained ETSEF model is evaluated on five tasks that belong to different domains, as they cover a wide range of medical imaging techniques and distribution shifts. For the first four domains, we utilised a source dataset of ImageNet, an intermediate source dataset from the medical area. The last domain utilised an out-of-distribution target dataset to test the ETSEF model's generalisation ability towards unseen data, and there is no source data in this task.}
\label{Overview of datasets}
\end{figure}

Fig. \ref{Overview of datasets} gives an overview of the datasets used for each task. The first four tasks represent common clinical scenarios involving training models for specific modalities and tasks, while the final setting simulates an extreme situation where a domain-specific pre-trained model for a rare disease or an unseen task is unavailable. To examine the robustness of ETSEF when faced with various clinical data challenges, all selected target data sets are limited in data size: the colourful training sets contain up to 6,000 images, and the greyscale training sets have up to 2,800 samples. Another challenge is data imbalance, as most of the target datasets have imbalanced data samples between classes. Other challenges, such as low resolution, proportional distortion, and variations in image characteristics, further complicate the task of learning powerful representations from these data. This created a very challenging environment for us to test the ETSEF model.

\subsection{Performance of ETSEF}

This section evaluates the ETSEF model using several metrics, including accuracy, Recall, Precision, and F1-score. Additionally, we compare our method with strong supervised baseline models, along with several state-of-the-art methods to demonstrate its performance.

\textbf{Evaluation Metrics.} After the training phase, all models were evaluated using test sets derived from the target datasets, employing the accuracy, recall, precision, and F1 score evaluation metrics to assess their performance on each task. These metrics are derived directly from the confusion matrix, which includes true positives (TP), true negatives (TN), false positives (FP), and false negatives (FN). A TP represents a positive event correctly predicted as positive, while a TN indicates a negative outcome correctly predicted. FP occurs when a negative outcome is incorrectly predicted as positive, and FN occurs when a positive outcome is incorrectly predicted as negative. The evaluation metrics of accuracy, recall, precision, and F1 score are all based on these four measurements and the equations for these metrics are listed below.

\begin{small}\centering
$$ \textbf{Accuracy} = (TP + TN)/ (TP + TN + FN + FP)$$
$$ \textbf{Recall} = TP/ (TP + FN)$$
$$ \textbf{Precision} = TP/ (TP + FP)$$
$$ \textbf{F1 Score} = 2 * \frac{Precision * Recall}{Precision + Recall}$$
\end{small}

Among these evaluation metrics, accuracy indicates the proportion of correctly predicted samples among all test samples. Recall (sensitivity) measures the proportion of actual positive samples correctly identified. Precision represents the proportion of true positive predictions among all positive predictions made by the model. The F1 score is a harmonic mean of recall and precision which provides a balanced metric that considers both precision and recall. 

\begin{figure}[htbp]
\centerline{\includegraphics[width=0.98\textwidth]{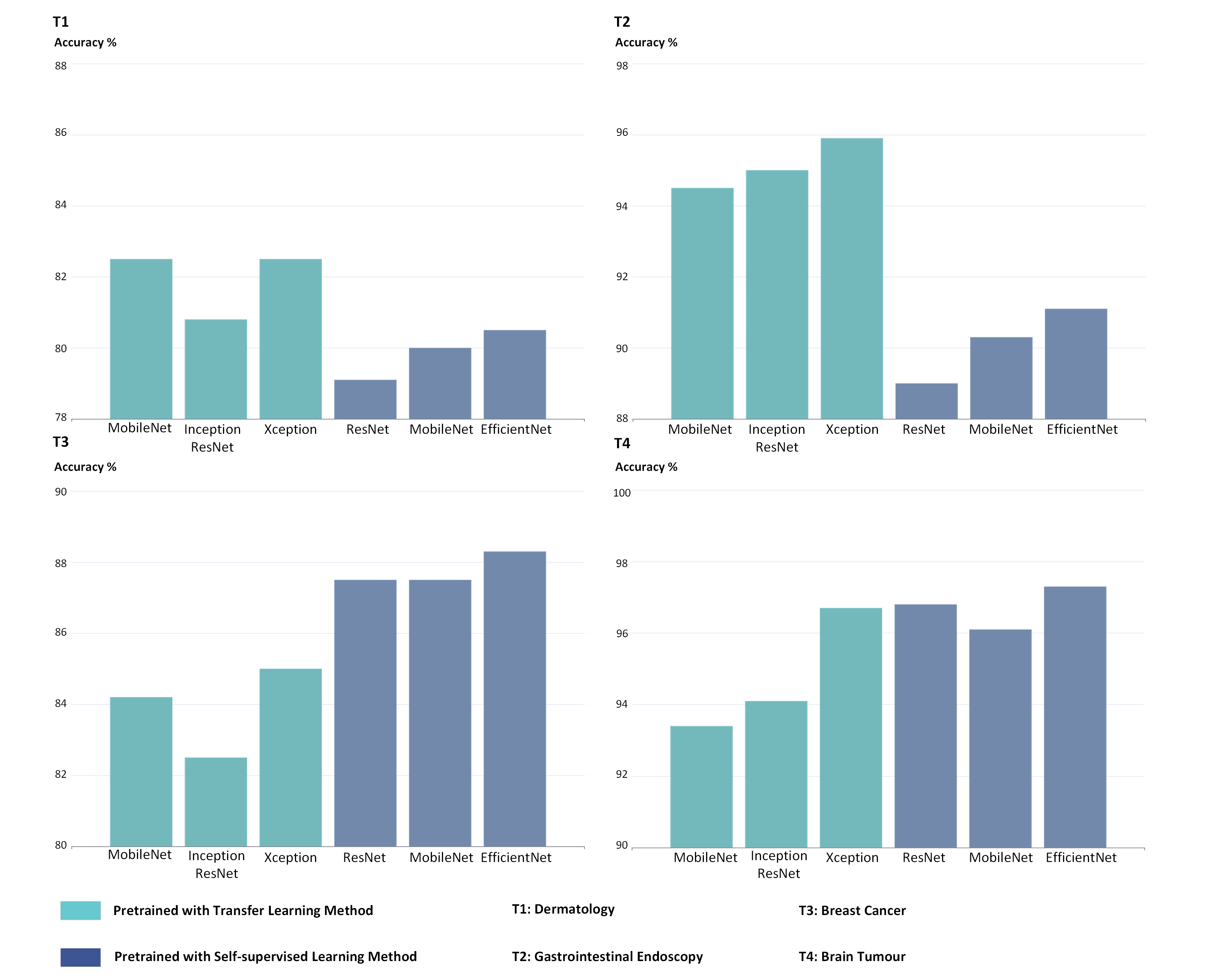 }}
\caption{\textbf{Performance of Pre-trained Base Models for Each Task:} The performance of the base models is evaluated by accuracy performance. Models pre-trained using TL methods are highlighted in green, while those pre-trained using SSL methods are marked in blue. Overview results show that pre-trained models using supervised TL methods perform better in target datasets with colourful images, and the models pre-trained with self-supervised methods are generally stronger in grey-scale image datasets.}
\label{Pre-trained model performance}
\end{figure}

\textbf{Performance on Domain-specific Tasks.} Fig. \ref{Pre-trained model performance} provides an overview of the pre-trained base models' accuracy performance on Task 1-4. Based on the figure, TL-based pre-trained models showed up to 7\% higher accuracy in Dermatology MonkeyPox classification($T_1$) and Gastrointestinal Endoscopy classification ($T_2$). In contrast, SSL-based models outperformed TL-based models in Breast Ultrasound Classification ($T_3$) and in Brain Tumour classification ($T_4$). The primary reason that causes the performance difference between the target groups $T_1$, $T_2$ and $T_3$, $T_4$ lies in the colour channels: $T_1$ and $T_2$ consist of colourful medical images, while $T_3$ and $T_4$ are grey-scale medical images. Notably, we minimised the influence of colourful ImageNet pre-trained weights by using similar-scale intermediate medical datasets with the same colour channels to retrain all base models during the representation learning stage. The results indicated that models using TL methods are more effective at learning from colourful images compared to SSL methods. Meanwhile, SSL methods demonstrated superior performance when learning from grey-scale images, which has been noticed by previous research but did not raise enough attention \cite{xie2018pre, ericsson2021well}. These observations suggest that both supervised and unsupervised pre-training methods have distinct advantages depending on the data environment. To achieve better performance, it is beneficial to consider both methods in combination rather than relying on a single approach.

\begin{figure}[htbp]
\centerline{\includegraphics[width=0.98\textwidth]{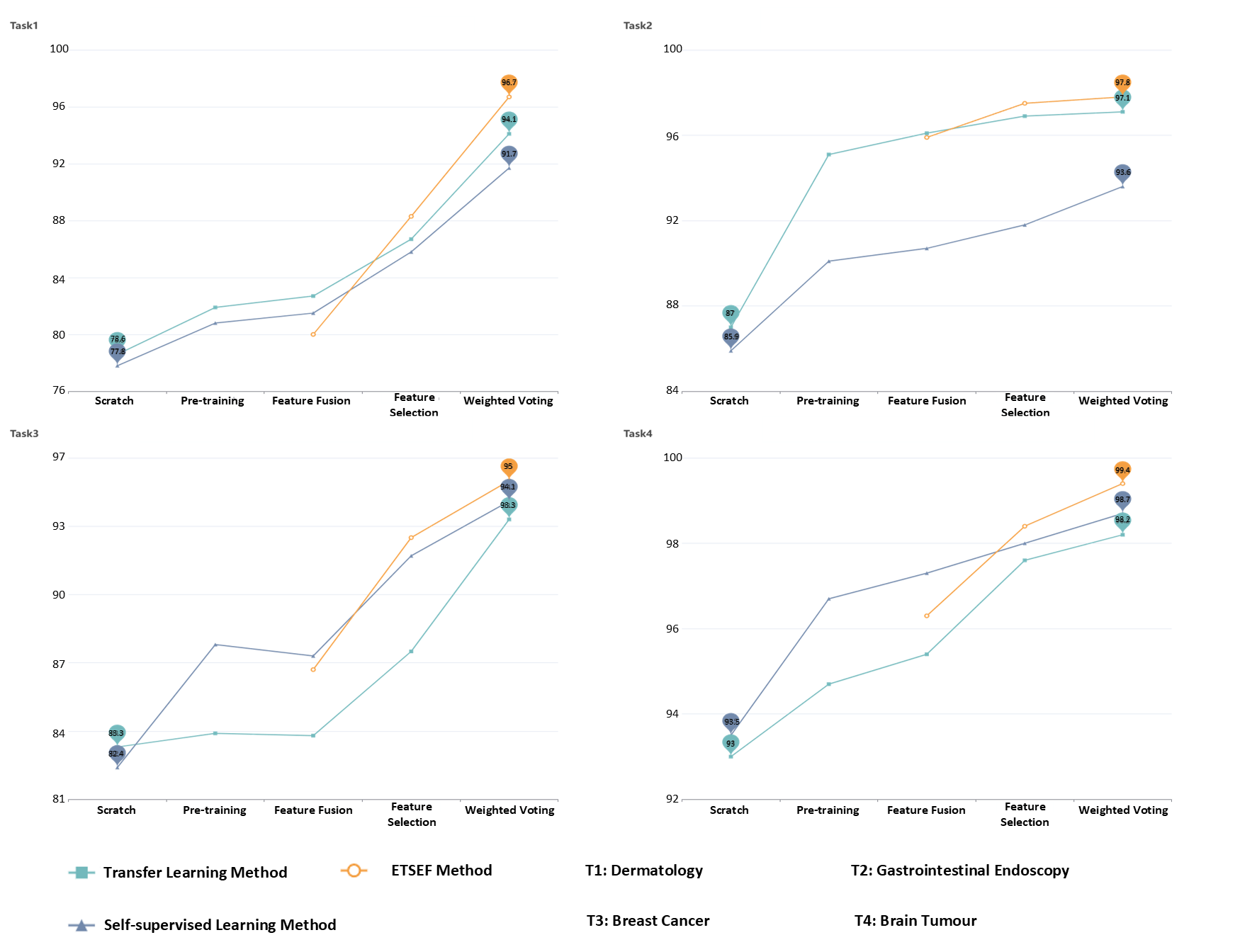}}
\caption{\textbf{Ensemble models' five-stage performance on each task:} This figure compares the performance of the baseline models and the ETSEF model across four main training stages: pre-training, feature fusion, feature selection, and majority voting. The results indicate a trend of performance improvement at each stage, benefiting both the baseline and ETSEF models. Notably, the ETSEF model outperforms the baseline models after the feature selection stage in every task.}
\label{Five-stage model performance}
\end{figure}

Fig. \ref{Five-stage model performance} shows the performance of each training stage during the ensemble learning process for the two baseline pre-training models and our ETSEF model. The performance comparison between the first and second stages indicated that using pre-trained learning techniques can generally enhance the model's overall performance. Among all the tasks, base models showed the most significant improvement in $T_2$ during this stage, likely due to its larger intermediate source dataset compared to the other tasks. However, the performance of the TL and SSL baseline models declined in $T_3$ after the feature fusion stage. This issue did not occur in other tasks, possibly because the features extracted from ultrasound medical images in $T_3$ can hardly distinguish each other and have a high chance of being redundant overlays. In the subsequent feature selection step, the ICA technique has also brought benefits to the ensemble model's performance, underscoring the importance of feature selection in maintaining the independence and generalisation ability of features. The final weighted voting stage further improved overall decision-making performance and ensured the robustness of the entire ensemble learning process.

Unlike the two baseline models, we hid the ETSEF model's early-stage performance and began comparison at the feature fusion stage, as it utilised the same TL and SSL pre-trained base models as the baseline models shown in the first two training stages. Initially, ETSEF's performance at the feature fusion stage was lower than that of the baseline ensemble models, possibly due to the same feature duplication issue. However, ETSEF outperformed both baseline models in each task after the feature selection and weighted voting stages. The results indicated that combining features from multiple pre-training methods and base model architectures leads to a statistically significant improvement compared to using a single pre-training method. The ETSEF model benefits from enhanced generalisation ability and a better capacity to capture useful representations. Notably, the performance of ETSEF showed greater improvement compared to the baselines in $T_1$ and $T_3$, which have smaller target dataset sizes, demonstrating the ETSEF method's effectiveness in scenarios with limited data samples.

\begin{table}[htbp]
\centering
\begin{adjustbox}{width=1\textwidth}
\begin{tabular}{  c c c c c c } 
\rowcolor{brown!40}
  \hline
   &  &  & & &  \\ 
  \rowcolor{brown!40}
 \multirow{-2}{4em}{\textbf{Tasks}}  & \multirow{-2}{4em}{\textbf{Method}} & \multirow{-2}{4em}{\textbf{Accuracy (\%)} }& \multirow{-2}{4em}{\textbf{Recall (\%)} } & \multirow{-2}{4em}{\textbf{Precision (\%)} } & \multirow{-2}{4em}{\textbf{F1score (\%)} }\\ 
  \hline

 \multirow{4}{4em}{Task 1} & Supervised Ensemble (Scratch) & 89.2 (88.7,89.7)& 89.2 (87.8,90.7)& 89.5 (89.3,89.7)& 88.9 \\ 

  & Supervised Ensemble(TL) & 94.2 (94.0,94.4)& 94.2 (94.0,94.4)& 95.2 (94.4,96.0)&94.3\\

  & Unsupervised Ensemble (SSL)& 91.7 (91.2,92.2) & 90.8 (90.5,91.1)&92.3 (91.9,92.7)&91.1\\

  & ETSEF (Proposed)& \textbf{96.7} (96.3,97.1)& \textbf{96.7} (96.2,97.2)& \textbf{96.9} (96.7,97.1)& \textbf{96.7}\\
  \hline
  
  \rowcolor{brown!30}
  & Supervised Ensemble (Scratch) & 89.3 (89.0,89.7)& 89.3 (89.1,89.5)& 89.2 (89.0,89.4)& 89\\ 

  \rowcolor{brown!30}
  & Supervised Ensemble(TL) & 97.1 (96.9,97.2)& 97.4 (97.2,97.5)& 97.4 (97.3,97.5)&97.4\\

  \rowcolor{brown!30}
  & Unsupervised Ensemble (SSL) & 93.4 (93.1,93.6) & 93.2 (93.1,93.3)&93.2 (93.0,93.3)&93.1\\

  \rowcolor{brown!30}
 \multirow{-4}{4em}{Task 2}  & ETSEF (Proposed)& \textbf{97.8} (97.5,98.0)& \textbf{97.8} (97.8,97.9)&\textbf{97.8} (97.8,97.8)&\textbf{97.8} \\
  \hline
  
  \multirow{4}{4em}{Task 3} & Supervised Ensemble (Scratch)& 87.5 (85.8,89.2)& 87.5 (87.2,87.8)& 87.5 (87.5,87.6)& 87.4 \\ 

  & Supervised Ensemble(TL) & 93.3 (93.0,93.6)& 90.8 (90.6,91.1)& 91.5 (91.0,92.0)&90.9\\

  & Unsupervised Ensemble (SSL) & 94.1 (94.0,94.2)& 93.3 (93.1,93.5)& 93.6 (93.2,94.0)&93.3\\

  & ETSEF (Proposed)& \textbf{95.0} (94.8,95.2)& \textbf{95.0} (95.0,95.0)& \textbf{95.0} (94.9,95.1)&\textbf{95.0}\\
  \hline
  
  \rowcolor{brown!30}
  & Supervised Ensemble (Scratch) & 95.7 (95.2,96.3)& 95.4 (93.7,97.1)& 95.7 (92.8,97.6)& 95.5\\ 

  \rowcolor{brown!30}
  & Supervised Ensemble(TL) & 98.2 (97.9,98.5)& 98.1 (98.0,98.2)& 98.3 (98.1,98.5)& 98.2\\

  \rowcolor{brown!30}
  & Unsupervised Ensemble (SSL) & 98.7 (98.3,99.1)& 98.9 (98.7,99.1)& 98.4 (97.7,99.1)&98.6 \\

  \rowcolor{brown!30}
  \multirow{-4}{4em}{Task 4}& ETSEF (Proposed)& \textbf{99.4} (99.2,99.6)& \textbf{99.4} (99.3,99.5)& \textbf{99.4} (99.1,99.5)& \textbf{99.4}\\
  \hline
\end{tabular}
\end{adjustbox}
\caption{\textbf{Comparison of ETSEF with baselines} The baseline performance consists of three ensemble scenarios: base models trained for target tasks without using pre-trained weights and models trained with TL or SSL pre-trained knowledge. The results are presented in terms of accuracy, recall, precision, and F1-score, with average performance reported as floating-point values. From the result comparison, ETSEF outperformed all baseline scenarios in every task, regardless of whether TL or SSL pre-trained methods were used.}
\label{ETSEF performance comparison}
\end{table}

Table. \ref{ETSEF performance comparison} presents a performance comparison between ETSEF and the three baseline models. While extended Fig. \ref{Confusion matrix baseline TL}, Fig. \ref{Confusion matrix baseline SSL}, and Fig. \ref{Confusion matrix} provide a corresponding confusion matrix based on the ETSEF model and two pre-trained ensemble baselines. According to the test performance, the ETSEF method showed substantial improvements compared to all baseline models, whether trained from scratch or with pre-trained weights. In particular, for dermatology, endoscopy, and breast cancer tasks, ETSEF's performance increased by 7.5\% to 8.5\% across all evaluation metrics when compared to the scratch baseline model. The improvements were even more pronounced in the brain tumour task, with an 11.3\% increase in accuracy and recall. Furthermore, ETSEF showed a significant improvement of up to 2.9\% in accuracy for each task compared to the two strong pre-trained ensemble baseline models. The experimental results demonstrated the effectiveness of combining multiple pre-training methods, as the combination method enhanced overall performance by expanding the base model architectures and pre-training method variety, thereby improving their generalisation ability on target datasets without requiring large amounts of training data. Compared with Fig. \ref{Confusion matrix} with Fig. \ref{Confusion matrix baseline}, the confusion matrix of the baseline ensemble model trained without data augmentation showed less balanced performance across different classes. This experimental finding suggests that the data imbalance issue observed in $T_1$, $T_3$, and $T_4$ has been well addressed and did not affect ETSEF's final performance. Highlighting the robustness of our approach when dealing with imbalanced medical datasets, which is particularly valuable in medical imaging tasks.

\textbf{Performance on Out-of-distribution Domain Task.} The experimental results of the ETSEF model and baseline models in the OOD data detection task are presented in Table. \ref{OOD task performance comparison}. Specifically, we compared the performance of scratch-trained and pre-trained baseline models to assess their effectiveness in unseen data and provide a contrast to the ETSEF model. These baseline models are built from corresponding base models of $T_2$. Compared to previous domain-specific tasks, the performance on the OOD task showed a more significant rise in accuracy with an average of 9.5\% when applying pre-trained weights to base models. This finding suggests that pre-trained methods can significantly enhance a model's capability to capture useful features when encountering new data with different feature distributions. Additionally, we noticed that the SSL baseline model outperformed the TL baseline model in this task, which contrasts with the previous observation that the TL method has better performance on colourful datasets in domain-specific tasks. This suggests that the features captured from the SSL method may cover a broader space, thus supporting better generalisation ability.

\begin{table}[htbp]
\centering
\begin{adjustbox}{width=1\textwidth}
\begin{tabular}{  c c c c c c } 
\rowcolor{brown!40}
  \hline
   &  &  & & &  \\ 
  \rowcolor{brown!40}
 \multirow{-2}{4em}{\textbf{Tasks}}  & \multirow{-2}{4em}{\textbf{Method}} & \multirow{-2}{4em}{\textbf{Accuracy (\%)} }& \multirow{-2}{4em}{\textbf{Recall (\%)} } & \multirow{-2}{4em}{\textbf{Precision (\%)} } & \multirow{-2}{4em}{\textbf{F1score (\%)} }\\ 
  \hline

 \multirow{4}{4em}{Task 5} & Supervised Ensemble (Scratch) & 63.2 (61.8,64.5)& 63.2 (63.0,63.4)& 63.2 (62.8,63.6)& 63.2 \\ 

  & Supervised Ensemble(TL) & 71.8 (71.2,72.4)& 71.8 (71.3,72.3)& 71.8 (71.4,72.2)&71.8\\

  & Unsupervised Ensemble (SSL)& 73.6 (73.2,73.9) & 73.6 (73.6,73.6)&73.7 (73.4,74.0)&73.6\\

  & ETSEF (Proposed)& \textbf{75.6} (74.7,76.3)& \textbf{75.6} (74.7,76.4)& \textbf{75.7} (95.3,96.0)& \textbf{75.6}\\
  \hline
  
  \hline
\end{tabular}
\end{adjustbox}
\caption{\textbf{OOD task performance comparison} The performance table indicates that the baseline ensemble model without any pre-training methods significantly underperforms compared to models utilising pre-training techniques. In contrast, the ETSEF model, which combines multiple pre-training methods has demonstrates superior performance when handling unseen data compared to the two baseline models that use a single pre-training method.}
\label{OOD task performance comparison}
\end{table}

Similar to the domain-specific tasks, ETSEF achieved the highest performance compared to all baselines. There was a substantial improvement of 12.4\% from the scratch baseline model to our ETSEF model and a 2\% improvement in accuracy compared to the pre-trained baseline models. The performance improvement demonstrated that ETSEF not only excels in domain-specific tasks but also handles unseen data more efficiently. Based on all these observations and analyses, we learned that pre-training methods are necessary for ensemble learning, which brings significant benefits to the final model. The superior performance of ETSEF further suggests that expanding the range of pre-trained knowledge can yield even greater advantages.

\textbf{Performance Comparison with State-Of-The-Art Methods.} To further demonstrate the performance and efficiency of ETSEF, we compare it with recent state-of-the-art models. Table \ref{Performance comparison with SOTA models} lists the performance of selected methods alongside our ETSEF method. We selected reference methods based on their publication year and citation count, focusing on research papers published within the recent three years to ensure their relevance. Pre-print methods without citations were excluded to maintain the trustworthiness of the comparisons. Notably, this comparison is conducted only within domain-specific tasks, as the main focus of the OOD task is to test the generalisation ability of the ETSEF model instead of achieving higher performance.

\begin{table}[htbp]
\centering
\begin{adjustbox}{width=1\textwidth}
\begin{tabular}{  c c c c c c c } 
\rowcolor{brown!40}
  \hline
   &  &  & & & & \\ 
  \rowcolor{brown!40}
 \multirow{-2}{4em}{\textbf{Tasks}}  & \multirow{-2}{4em}{\textbf{Method}} & \multirow{-2}{4em}{\textbf{Model}} & \multirow{-2}{4em}{\textbf{Accuracy (\%)} }& \multirow{-2}{4em}{\textbf{Recall (\%)} } & \multirow{-2}{4em}{\textbf{Precision (\%)} } & \multirow{-2}{4em}{\textbf{F1score (\%)} }\\ 
  \hline

 \multirow{4}{4em}{Task 1} & Ali et al. (2023) \cite{ali2023web}& DenseNet121 & 82.3 & 80.0 & 85.0 & 83.0 \\ 

  & Naysk et al. (2024) \cite{nayak2024mpox}& Mpox Classifier& 94.0 & 93.0 & 93.5 & 93.0 \\

  & Biswas et al. (2024) \cite{biswas2024large}& BinaryDNet53 & 95.0 &  94.4 &  91.5 & 92.9\\

  & ETSEF (Proposed)&CNN Ensemble  &\textbf{96.7}& \textbf{96.7} & \textbf{96.7} & \textbf{96.7}\\
  \hline
  
  \rowcolor{brown!30}
  & Dong et al. (2023) \cite{dong2023local} & Transformer Ensemble & 90.6 & - & - & 90.7\\ 

  \rowcolor{brown!30}
  & Mukhtorov et al. (2023) \cite{mukhtorov2023endoscopic} & ResNet-152 &93.5 & - & - & - \\

  \rowcolor{brown!30} 
  & Wang et al. (2023) \cite{wang2023vision} & VIT & 95.4 & - & - & 95.4\\

  \rowcolor{brown!30} 
  & Patel et al. (2024) \cite{patel2024classification} & EfficientNetB5 & 92.6 & 93.0 & 93.0 & 93.0 \\

  \rowcolor{brown!30}
   & Khan et al. (2024) \cite{khan2024multi} & LG-CNN & 95.0 & 95.0 & 95.0 & 95.0\\

  \rowcolor{brown!30}
 \multirow{-6}{4em}{Task 2}  & ETSEF (Proposed)& CNN Ensemble &\textbf{97.8} & \textbf{97.8} &\textbf{97.8} &\textbf{97.8} \\
  \hline
  
  \multirow{6}{4em}{Task 3} & Gheglati et al. (2022) \cite{gheflati2022vision} & VIT & 85.3 & - & - & - \\ 

  & Deb et al. (2023) \cite{deb2023breast} & CNN Ensemble & 85.2 & - & - & -\\

  & Sahu et al. (2023) \cite{sahu2023high} & Hybrid CNN & 93.7 & 94.6 & 93.2 & -\\

  & Yue et al. (2024) \cite{yue2024medmamba} & MedMamba & 87.2 & - & - & - \\

  & Zhou et al. (2024) \cite{zhou2024distilling} & VIT-Ensemble & 93.8 & 91.7 & 95.0 & - \\

  & ETSEF (Proposed) & CNN Ensemble &\textbf{95.0} & \textbf{95.0}& \textbf{95.0} &\textbf{95.0}\\
  \hline
  
  \rowcolor{brown!30}
  & Kadam et al. (2021)\cite{kadambrain2021brain} & CNN & 94.0 & - & - & 94.0\\ 

  \rowcolor{brown!30}
  & Ravinder et al. (2023) \cite{ravinder2023enhanced}& GCN & 95.0 & - & - & -\\

  \rowcolor{brown!30}
  & Sahu et al. (2024) \cite{sahu2024deep} & CNN & 94.2 & 93.5 & 93.5 & 93.5\\

  \rowcolor{brown!30}
  & Sachdeva et al. (2024) \cite{sachdeva2024comparative}& Xception & 97.4 & - & - & 94.9\\

  \rowcolor{brown!30}
  \multirow{-5}{4em}{Task 4}& ETSEF (Proposed)&CNN Ensemble  &\textbf{99.4} & \textbf{99.4} & \textbf{99.4} & \textbf{99.4}\\
  \hline
\end{tabular}
\end{adjustbox}
\caption{\textbf{Performance comparison with state-of-the-art models} We compared the ETSEF model with recent research using the same target datasets. The results show that the proposed ETSEF framework demonstrates advanced performance across various medical imaging domains and superior capability in handling challenging multi-classification tasks.}
\label{Performance comparison with SOTA models}
\end{table}

\textbf{ETSEF show superior performance when compared to state-of-the-art methods.} For $T_1$, the proposed ETSEF method outperformed all reference methods, showing an accuracy advantage ranging from 1.7\% to 14.5\%. Although Nayak et al. \cite{nayak2024mpox} and Biswas et al. \cite{biswas2024large} achieved a 10\% performance advantage over our pre-trained base models with specifically designed architectures for the MSLDv2 dataset, the ETSEF method leverages an ensemble framework to enhance the base models' performance significantly, and finally surpassing these specialised methods.

We observed numerous related research efforts in the gastrointestinal endoscopy classification task ($T_2$). The listed methods include popular Convolutional Neural Network (CNN) architectures like ResNet and EfficientNet, as well as transformer-based models. Compared to the works of Mukhtorov \cite{mukhtorov2023endoscopic} and Patel \cite{patel2024classification}, our base models which utilised similar CNN architectures with fewer layers have achieved higher performance through the double fine-tuning steps. Combining these powerful base models with the ensemble learning technique resulted in a 2.5\% improvement over the highest reference method, which used the advanced vision transformer model \cite{wang2023vision}.

For the breast cancer classification task, we noticed that more research is being done to employ ensemble or hybrid methods to enhance model performance. This is possibly due to the limited sample size of the target BusI dataset that has constrained the single DL model's performance. We observed that methods with ensemble learning generally performed better than single-model approaches. The key difference between our method and these referenced ensemble methods is that we combined multiple base model architectures and utilised various pre-training methods simultaneously to learn more generalised representations from the source domain. The results highlight the efficiency of our method in learning useful knowledge.

In the brain tumour classification task, the reference methods that used different CNN models achieved a high average performance, with the highest accuracy being 97.4\%. In contrast, our ETSEF method achieved a perfect accuracy of 99.4\% in test cases, demonstrating that the ETSEF method can correctly classify almost all samples across four classes, offering a significant performance advantage over advanced methods.

\subsection{Ablation Study}

To understand the importance and contributions of different base models in the ensemble learning stage, we conducted an ablation study to analyse their impact on the higher-level ML classifiers.

\begin{figure}[htbp]
\centerline{\includegraphics[width=0.98\textwidth]{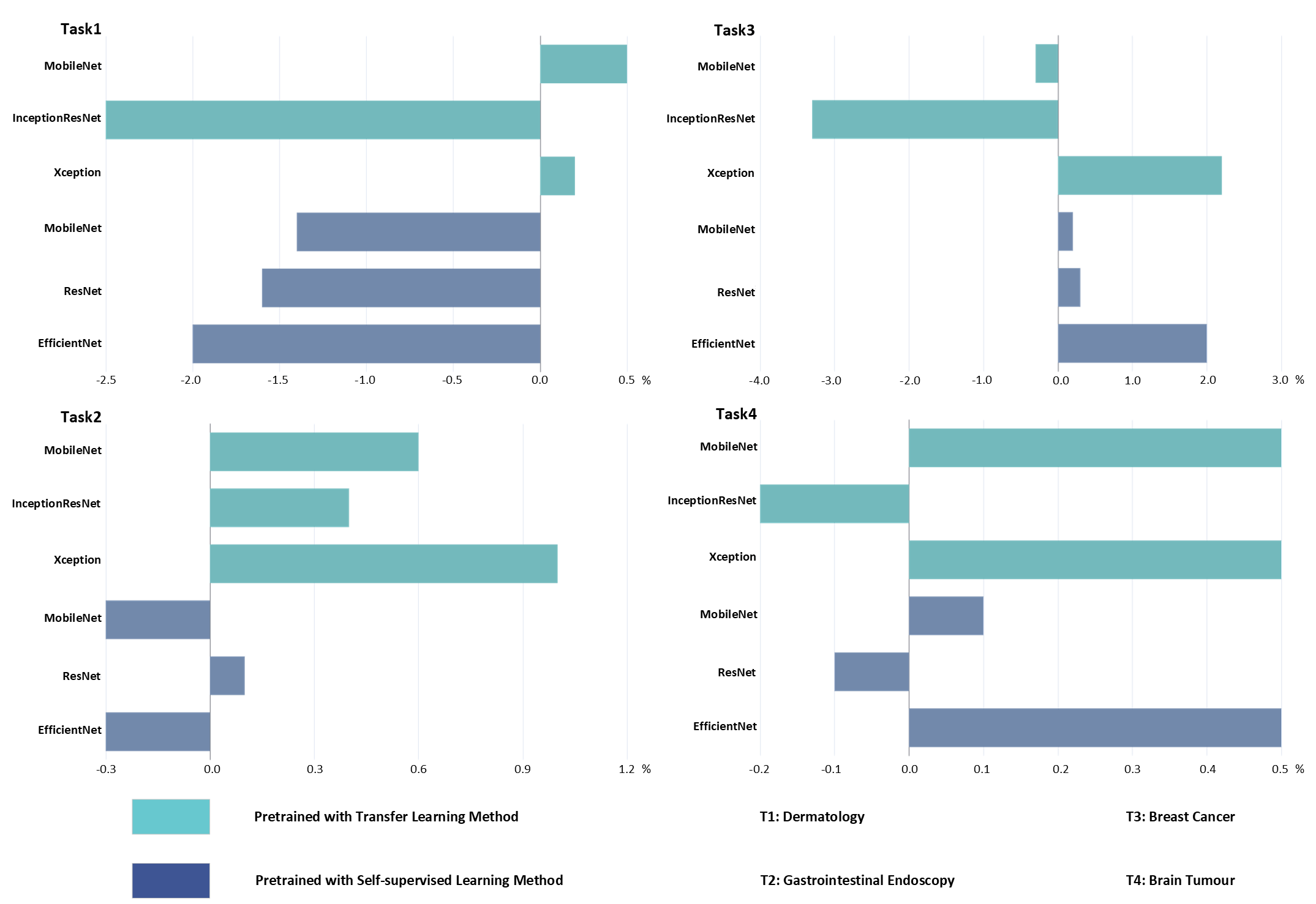}}
\caption{\textbf{Ablation study of pre-trained base models:} An ablation study was conducted to understand the influence of each base model in the ensemble learning stage, providing evidence of their value to the overall model performance. The results indicate that TL base models generally have a positive influence on tasks 1 and 2, while SSL base models are more effective in supporting ensemble learning on tasks 3 and 4.}
\label{Ablation study}
\end{figure}

We set up the ablation study by excluding one base model at a time and using the remaining models to conduct ensemble learning. We compared the performance of these modified models with the ETSEF model that included all base models to assess the influence of each base model. Specifically, we extracted features from each base model, then concatenated and selected them using the same methods as in the ETSEF model. We trained these processed features with the same higher-level ML classifiers (SVM, XGBoost, RF, NB, and KNN) as in the previous experiment. Finally, we averaged the performance of each ML classifier to obtain reliable results and compared these averages to determine the impact of excluding each base model on the ETSEF model's performance.

Fig. \ref{Ablation study} illustrates the performance changes observed in each task, highlighting the impact of each base model on the ML classifiers' performance. Notably, if excluding a base model decreases the ML classifiers' performance, it indicates that the chosen base model is necessary for the ensemble learning process and improved integrated feature quality. Conversely, if excluding a base model enhances the ML classifiers' performance, it suggests that including the base model negatively impacts the ML classifiers' input feature quality.

The figure also reveals that the influence of different pre-training methods varies across different tasks. For example, the MobileNet model pre-trained with TL positively influences tasks 1, 2, and 4 but negatively impacts task 3. In contrast, the same MobileNet model pre-trained with SSL positively influences tasks 3 and 4 but negatively affects tasks 1 and 2. Generally, TL-based models positively influence tasks involving colourful medical image datasets (tasks 1-2), while SSL-based models positively influence tasks with grey-scale datasets (tasks 3-4). This observation aligns with the performance trend from Fig. \ref{Pre-trained model performance}: TL-based models perform better with colourful datasets, providing higher-quality features, and SSL-based models perform better with grey-scale datasets.

However, this general observation does not cover all base models, as the Xception model pre-trained with TL consistently shows a positive influence, while the SSL-based ResNet model can also positively impact colourful datasets (task 2), and the TL-based InceptionResNet model has a negative influence on task 1. These findings suggest that the final performance of base models depends not only on the pre-training method and architecture but also on the characteristics of the target dataset and many other factors. The influence between different architectures, training methods, and data environments remains unresearched, waiting for more exploration in the future. At this stage, our ablation study mainly indicates that choosing suitable pre-training methods for different target datasets may maximise the positive impact on ensemble learning. Additionally, incorporating diverse base model architectures and pre-training methods can continue to enhance the ensemble model's ability to learn and extract high-quality representation features and achieve more stable performance across varying environments.

\subsection{Visual Analysis}

Medical image analysis involves high-stakes decision-making that is traditionally reliant on human domain experts, where the decision-making process can be described and understood. However, neural networks, consisting of numerous layers contributing to the final predictions, present a challenge in fully comprehending their decision-making process \cite{van2022explainable}. To test the robustness of the trained ETSEF model and understand how the model makes predictions, we employed several explainable artificial intelligence (XAI) techniques, including Grad-CAM, SHAP, and t-SNE. These XAI techniques allow us to visualise the attention areas of the trained models that support their decision-making and examine the extracted feature distributions to understand how the model classifies the features. By comparing the attention areas of the base models with the ground truth areas that indicate disease, we can evaluate how well the models have learned to recognise the true disease symptoms. Additionally, this helps us determine whether the model's predictions are based on reliable judgments or influenced by unrelated elements.

\textbf{Deep SHAP map analysis.} The SHAP method \cite{ribeiro2016should} calculates the marginal contribution of each feature by comparing the model's prediction with and without the feature. This process involves masking the feature, observing the change in the model's output, and aggregating the contributions using Shapley values. Specifically, a feature point in the SHAP map with a higher SHAP value indicates a greater contribution to the model's final prediction, while a lower SHAP value means it detracts from the model's decision for the corresponding label.

\begin{figure}[htbp]
\centerline{\includegraphics[width=0.9\textwidth, height=17cm]{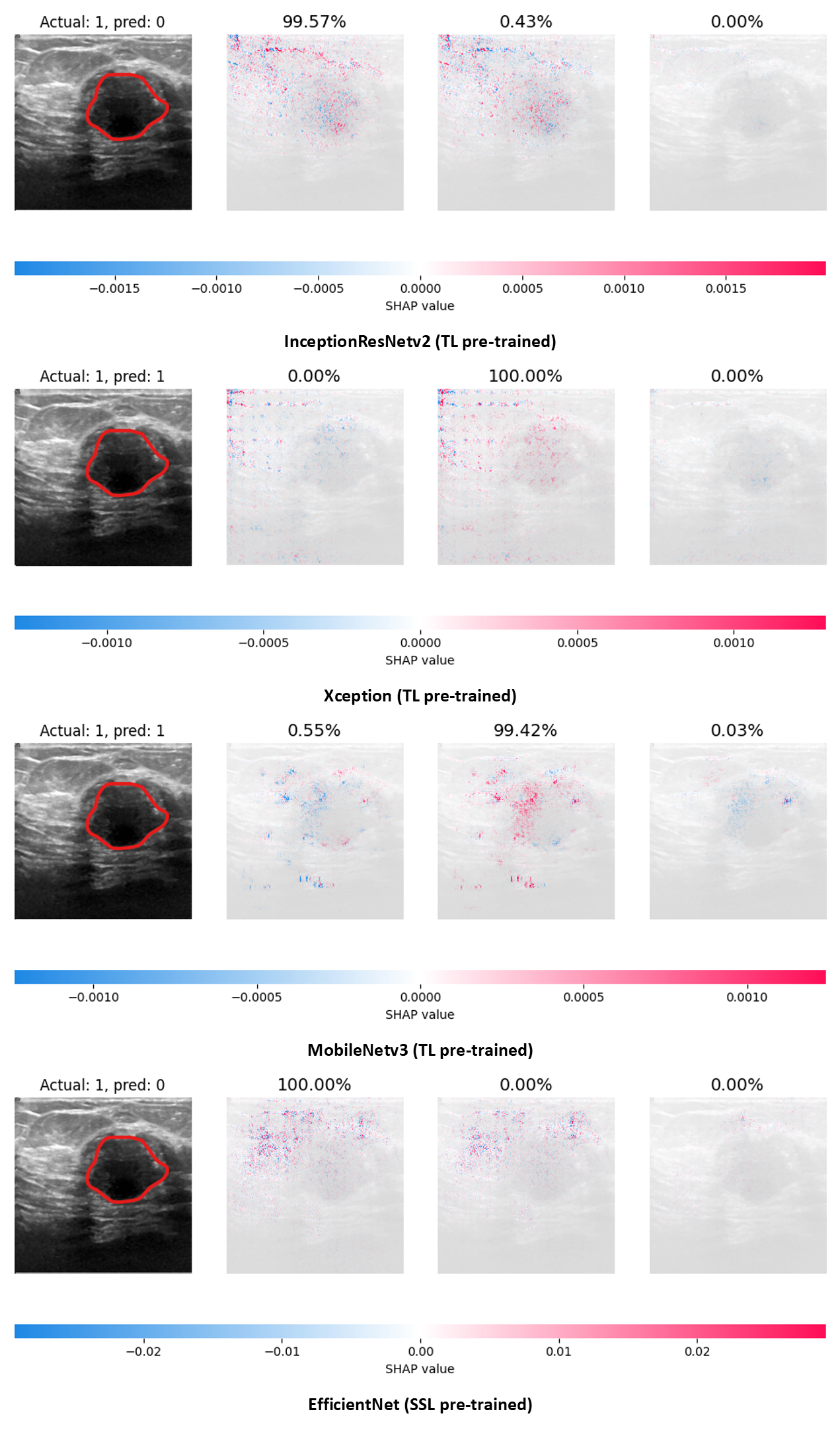}}
\caption{\textbf{Explainable visualisation of base models using SHAP maps:} The SHAP explainable AI technique is used to visualise the base model's attention areas and captured representations in the target image samples. The maps demonstrate that when pre-trained SSL base models occasionally fail to capture useful disease representations from the target sample, pre-trained TL base models can provide additional support.}
\label{SHAP map task3}
\end{figure}

As mentioned previously in Fig. \ref{Pre-trained model performance}, SSL pre-trained base models generally outperform TL pre-trained models in grey-scale medical imaging tasks, whereas TL models excel with colourful medical images. However, it is premature to conclude that one pre-training method is superior to the other under specific conditions, as performance metrics alone do not fully reflect how well a model has learned from a dataset. Therefore, we compare the decision-making processes of each base model using deep SHAP maps to assess their robustness with the target dataset. We selected the EfficientNet model, which performed best among the SSL base models, and compared it with the three TL base models in the grey-scale imaging environment using the BUSI dataset from $T_3$. Fig. \ref{SHAP map task3} provides a sample group of SHAP maps generated from four pre-trained base models, each containing four target image sub-figures. The first column shows the original test sample image, along with the true label and the predicted label made by each base model. The target image is a sample of malignant breast cancer using ultrasound imaging, where the symptom area appears as a shadow hole in the middle of the image. The remaining three columns represent all the classes in the target dataset, with the percentage number above each sub-figure indicating the probability that each model assigns to classifying the test image into the corresponding label. 

From the SHAP maps, we observed that the EfficientNet model incorrectly predicted the target image as benign, focusing its attention on the top left of the image. Similarly, the InceptionResNet model, which pre-trained using the TL method, made the same incorrect prediction based on the middle part of the shadow hole and the top left area. In contrast, the TL-based Xception and MobileNet models correctly classified the target sample as malignant. The Xception model distributed its attention between the middle of the ground truth shadow area and the top left of the image, while the MobileNet model concentrated more on the shadow hole itself, covering the left half of the ground truth disease area.

\begin{figure}[htbp]
\centerline{\includegraphics[width=0.9\textwidth, height=17cm]{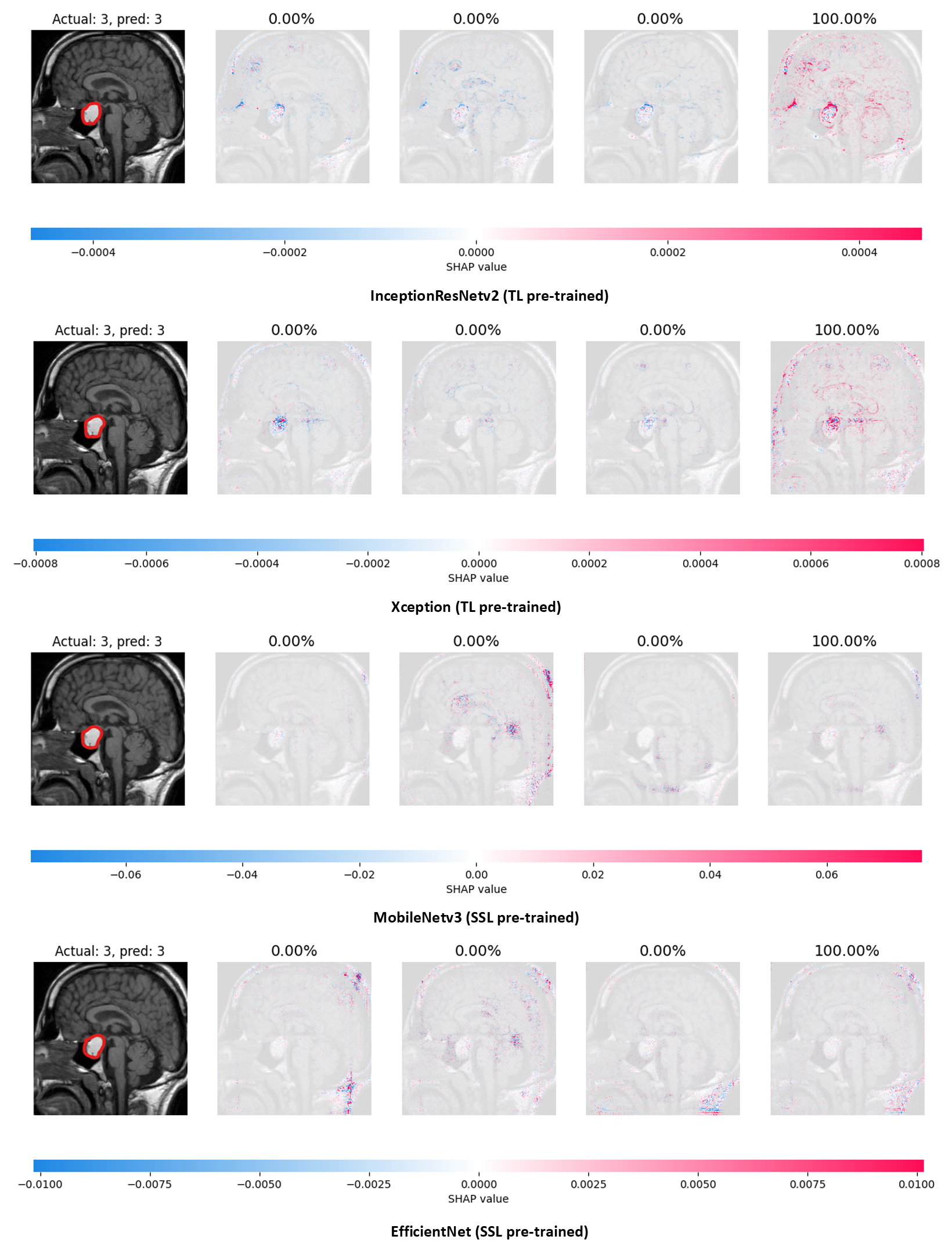}}
\caption{\textbf{Explainable visualisation of base models using SHAP maps:} Here we provide another example where SSL base models fail to capture the representative disease symptoms. In this case, TL-based models offer a better understanding of the target disease symptoms, compensating for the SSL models' lack of knowledge.}
\label{SHAP map task4}
\end{figure}

Fig. \ref{SHAP map task4} provides a visualisation sample from $T_4$. The ground truth label for the sample image is a pituitary tumour caused by unusual growths in the pituitary gland. In the original image, the pituitary tumour symptoms are marked with a red circle located behind the nose at the base of the brain. For this case, we compared two heatmaps generated from the TL base models (InceptionResNet and Xception) with two heatmaps from SSL base models (MobileNet and EfficientNet). The SHAP maps show that all four pre-trained base models made correct predictions. However, the feature map shows that the TL-based models made their predictions by capturing the ground truth symptoms within the pituitary gland area. In contrast, the SSL-based models focused on the right bottom and meninges areas, making their predictions less convincing and reliable compared to the TL-based models.

To summarise, we observed samples suggesting that SSL pre-trained base models do not always learn better symptoms than TL-based models under a grey-scale image environment, and TL models can provide useful insights when SSL models fail. Also, a similar situation happened when pre-trained base models faced the colourful image environment as the extended Fig. \ref{SHAP map task1} and Fig. \ref{SHAP map task2} present. Importantly, from all these sample test cases, we noticed that none of the single pre-trained base models successfully captured the entire ground truth area, likely due to the limited size of the target dataset constraining the model's ability to learn high-quality representations and make reliable predictions. 

These observations have raised our attention as they pose a potential risk in real clinical settings, and they suggest that using a single deep learning architecture or pre-training method may be insufficient in limited and challenging medical environments. Fusing features from various base models shows promise in bridging this gap, as they combine the captured representation areas from each base model. Moreover, we observed more samples demonstrate that by combining different models' attention and extending the coverage, an ensemble model can make more reliable predictions. In the following subsection, we use Grad-CAM and t-SNE methods to provide further evidence of the power of feature fusion and ensemble learning in improving model robustness and reliability in the decision-making process.

\textbf{Grad-CAM heatmap analysis.} The Grad-CAM method used here is an advancement of the Class Activation Mapping (CAM) technique, which was originally developed by Zhou et al. \cite{zhou2016learning}. CAM integrates a global average pooling layer into a standard CNN, linking critical feature contributions to the CNN's specific predictions and providing insight into the model's decision-making process. Furthermore, Grad-CAM builds on this by using gradients from the network's final convolutional layer to produce a coarse localisation map. This map highlights the influential regions within an image that contribute to the prediction of specific concepts or classes. We used the Grad-CAM technique to generate heatmaps to pinpoint the critical area affecting the model's prediction.

\begin{figure}[htbp]
\centerline{\includegraphics[width=0.95\textwidth]{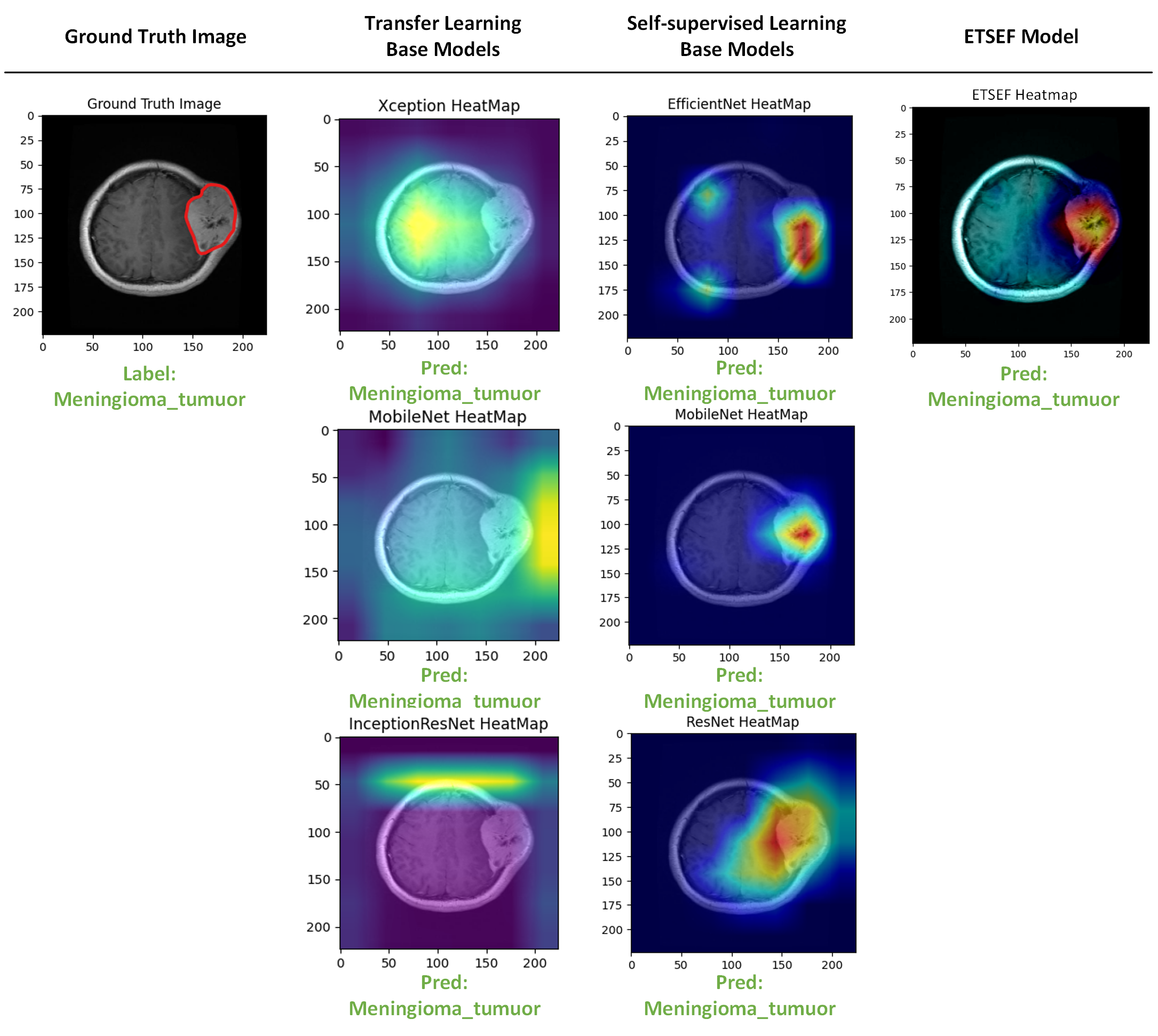}}
\caption{\textbf{Explainable visualisation using Grad-CAM maps:} To provide a comprehensive insight into the transformation of the model's decision-making process after ensemble training, we utilised Grad-CAM to map the models' attention regions. The sample heatmaps are from each pre-trained base model, compared with the ETSEF model that fused their features, and have clearly illustrated how the ensemble learning method improved the model's capability to capture target disease symptoms accurately.}
\label{GradCAM heatmap task4}
\end{figure}

Fig. \ref{GradCAM heatmap task4} shows a heatmap sample from $T_4$ of the BTC dataset. The target image is of a meningioma tumour, which is a common type of tumour that grows from the membranes surrounding the brain and spinal cord. The symptoms of this type of tumour include abnormal swelling that presses on the nearby brain, nerves, and vessels.

From Fig. \ref{GradCAM heatmap task4}, all pre-trained base models correctly predicted the ground truth label. However, TL-based models did not accurately capture the symptomatic area. Specifically, the Xception model focused on the left side of the brain, while the MobileNet and InceptionResNet models focused outside the brain. Conversely, the SSL-based models performed better, with all three identifying the abnormal area and concentrating on the ground truth region. The EfficientNet model partially covered the disease area, but also focused on two regions outside it. The MobileNet and ResNet models also partially captured the symptomatic area but did not fully reveal it.

The ETSEF model demonstrated a superior understanding of the target disease. Its heatmap perfectly captured the disease area and minimised the negative influence of the TL-based models, avoiding attention outside the brain.

\begin{figure}[htbp]
\centerline{\includegraphics[width=0.95\textwidth]{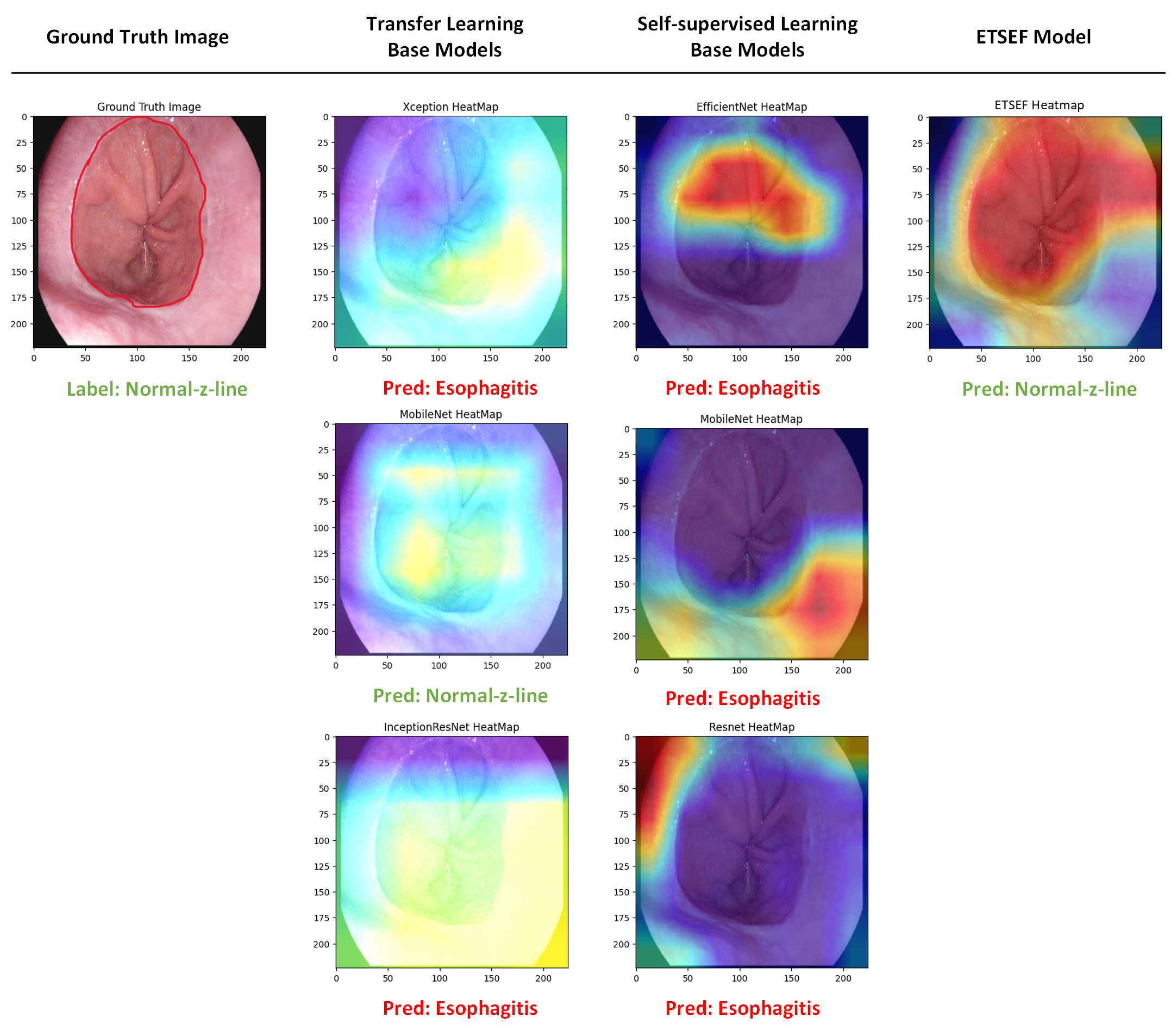}}
\caption{\textbf{Explainable visualisation using Grad-CAM maps:} Another representative sample shows that both TL and SSL-based models are unable to capture the target representation region fully. In contrast, the ETSEF model improves overall understanding and perfectly captures the disease symptom area.}
\label{GradCAM heatmap task2}
\end{figure}

Fig. \ref{GradCAM heatmap task2} compares Grad-CAM heatmaps between pre-trained base models and the ETSEF model using a sample image from the Kvasirv2 dataset of $T_2$. This image belongs to the normal Z-line class, where classification performance is often confused with esophagitis. The Z-line is the squamocolumnar junction where the esophageal squamous epithelium transitions to the stomach's columnar epithelium, while esophagitis indicates inflammation of the esophagus, characterised by an irregular, thickened Z-line with erythema, visible erosions, and ulcers.

In the figure, the first column shows the ground truth area marked on the original sample image, where the esophageal squamous epithelium is located in the middle of the image. The second column displays heatmaps from three TL-based pre-trained models. The Xception and InceptionResnet models only partially focused on the ground truth area and failed to make the correct prediction. In contrast, the MobileNet heatmap successfully covered most of the ground truth area and made the correct prediction. Self-supervised base models also struggled, with only the Efficient model successfully capturing the upper side of the ground truth area, while MobileNet and ResNet misfocused on areas outside the target.

The final column presents the heatmap of the ETSEF model after feature fusion. This model combined features from six base models, successfully capturing the target ground truth area and making an accurate prediction based on the enhanced features. The heatmap samples demonstrated that the ETSEF method extended the attention area of the target image and improved the feature quality to cover the whole ground truth area and make reasonable predictions.

Extended Fig. \ref{Extend GradCAM heatmap} presents additional samples from target tasks. These heatmaps demonstrate that the ETSEF model has enhanced the overall model's generalisation ability and reliability. It broadens the attention area when individual base models only capture parts of it or by correcting the focus when base models incorrectly focus on areas outside the disease symptoms.

\begin{figure}[htbp]
\centerline{\includegraphics[width=1\textwidth,height=17.6cm]{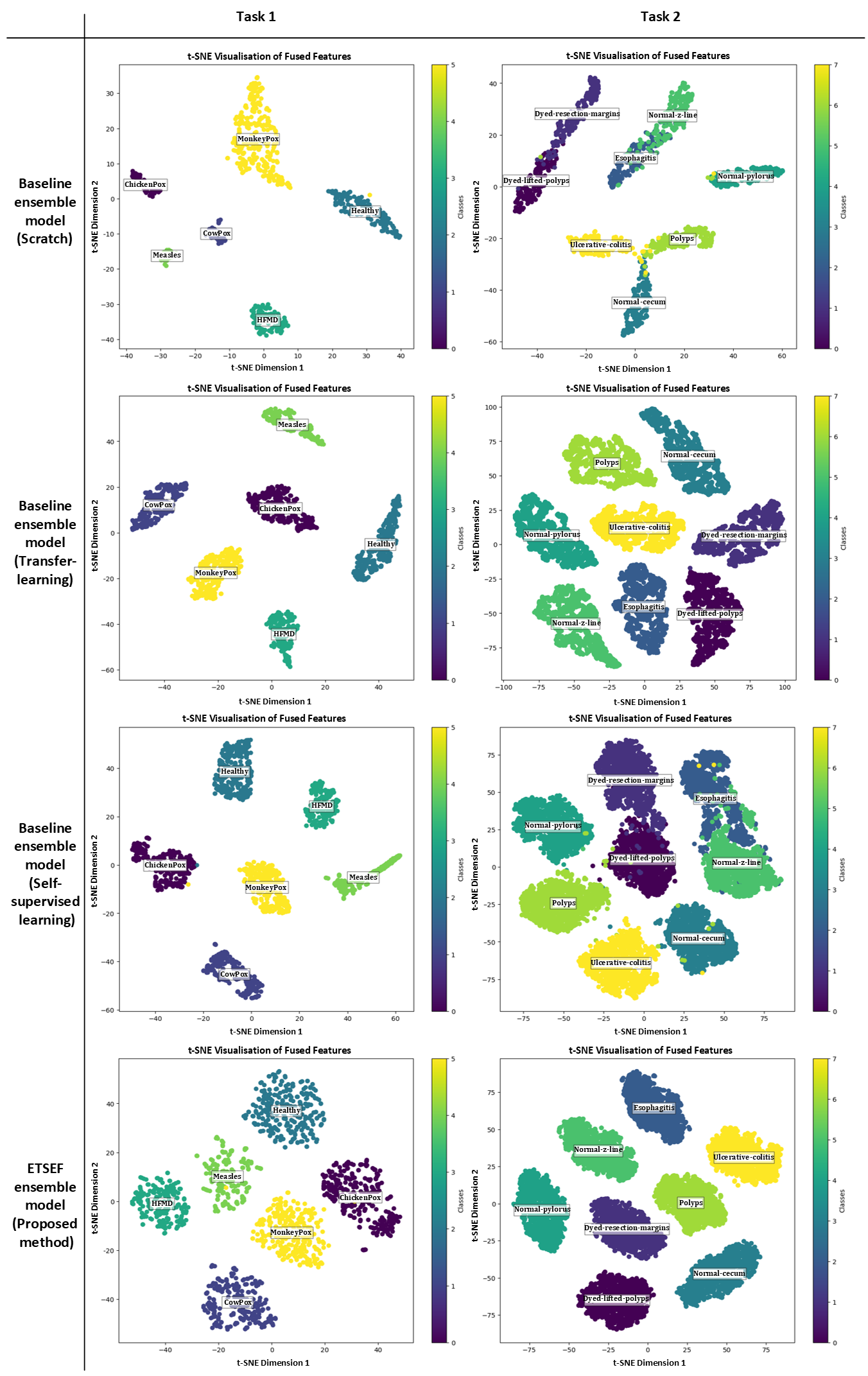}}
\caption{\textbf{Explainable visualisation using t-SNE maps:} To analyse the differences in feature quality after the feature fusion and selection steps, we used the t-SNE technique to visualise the processed feature distribution. We compared the feature classification performance of all baseline models with the ETSEF model. The feature map shows improved classification behaviour of the ETSEF method in $T_1$ and $T_2$.}
\label{t-SNE visualization map comparison1}
\end{figure}

\begin{figure}[htbp]
\centerline{\includegraphics[width=1\textwidth,height=17.8cm]{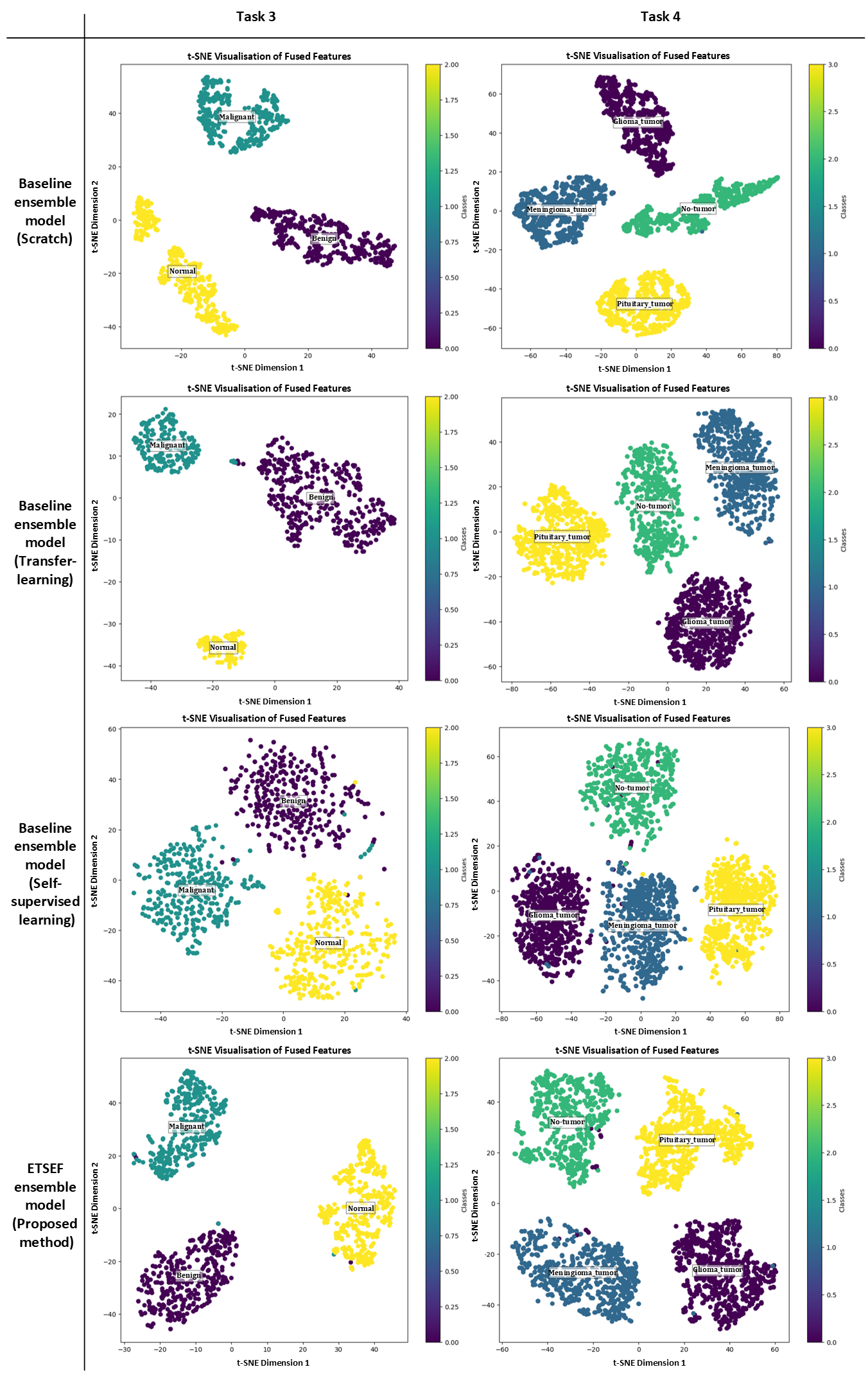}}
\caption{\textbf{Explainable visualisation using t-SNE maps:} This figure compares the baseline models with the ETSEF model using the t-SNE visualisation technique. The ETSEF method's feature map shows clearer boundaries in $T_3$ and $T_4$ when compared to the baselines that do not use multiple pre-training methods.}
\label{t-SNE visualization map comparison2}
\end{figure}

\textbf{t-SNE feature distribution analysis.} The t-SNE is a technique that visualises high-dimensional data by giving each data point a location in a two or three-dimensional map \cite{van2008visualizing}. Our experiment used the t-SNE technique to map each feature point at a two-dimensional map based on the feature point vectors. Fig. \ref{t-SNE visualization map comparison1} and Fig. \ref{t-SNE visualization map comparison2} provide a visualisation map that compares t-SNE maps between the ETSEF model and baseline ensemble models, where the baseline model's features have also undergone fusion and selection processes.

For task 1, the baseline model trained from scratch shows a biased feature distribution between classes. The yellow points representing the monkeypox class are less centralised than other classes, with some features mixed with the healthy class. The TL baseline model group features better than the SSL baseline, as the SSL baseline misclassified chickenpox features into monkeypox and healthy classes. In contrast, the ETSEF model's feature map is clearly classified, with distinct boundaries and similar-sized feature groups.

For Task 2, the baseline model trained from scratch produces a map with no clear boundaries, mixing features from all groups, making them hard to distinguish. The SSL baseline's feature quality is worse than the TL baseline, as the SSL feature map mixes features from esophagitis with normal-z-line class and dyed-resection-margins with dyed-lifted-polyps class. In contrast, the TL baseline classifies each class more distinctly, but the ETSEF model provides a more centralised and well-classified feature group, with clear boundaries and distances between each class. 

For Task 3, the baseline model from scratch poorly groups features, splitting malignant and normal class features into two subgroups with many features decentralised from the group centre. The TL baseline misclassified benign and malignant features, mixing some features between the two groups. At the same time, the SSL baseline has similar issues with features from benign and malignant classes mixed. The ETSEF model produces the best-classified feature map, with clear boundaries for all three groups and only a few feature points outside their groups.

For Task 4, all baseline and ETSEF models clearly define and classify feature groups according to their classes. These maps align with our observations from the performance table ~\ref{ETSEF performance comparison}, as all models achieved extremely high performance in this task. Still, the ETSEF method has a more centralized feature group with fewer outlier feature points compared to the TL and SSL baselines.

In summary, comparing the ETSEF model with three baseline models shows that our ETSEF strategy not only provides better model performance but also yields higher feature quality. This is a crucial finding because we have learned that a single pre-trained base model cannot fully understand and learn a satisfactory target representation with limited target data. While the ETSEF method can enhance the base model's generalisation and provide more robust predictions. The analysis of t-SNE feature maps demonstrates that the ETSEF method, which combines multiple DL architectures and pre-training methods, has better feature quality than standard ensemble methods using a single pre-training method and a few architectures. The combination of various pre-training methods and architectures makes the ETSEF method more robust and reliable in real-world clinical environments.

\section{Discussion}\label{sec3}

The challenge of data scarcity has long constrained the application of DL techniques in domains requiring specialised knowledge, such as medical, satellite, and industrial fields. Recently, advancements in computer hardware have provided large-scale computational capabilities, enabling the creation of more robust and powerful models using extensive data volumes. Additionally, the rise of self-supervised and semi-supervised approaches, which leverage unlabelled data at scale, has further driven this trend \cite{goyal2019scaling, rong2020self}. Researchers are increasingly focusing on modified large language models, such as vision transformers and other large-scale transformer-based architectures, in conjunction with self-supervised methods to address these issues \cite{caron2021emerging, chen2021empirical}. However, the high computational costs associated with these approaches often make them inaccessible to individual users, limiting their use within large companies and laboratories.

Previous studies to compare TL and SSL methods \cite{zhao2023comparison} reveal that TL methods offer advantages such as superior performance on colourful images, faster training speeds, and lower computational costs. These benefits make TL methods difficult to replace entirely with SSL approaches. Moreover, our ablation study suggests that the final performance of the base model varies with different DL architectures, making it hard to select the best choice under different circumstances. Especially when the ensemble learning process introduces more feature relationships and factors that may influence the final prediction. 

Based on these insights, we developed this generalist ensemble framework ETSEF, that leverages both TL and SSL methods to create a more efficient and robust model capable of performing well in challenging clinical environments with limited data samples. Experimental results presented in Fig. \ref{ETSEF performance comparison} and Fig. \ref{Performance comparison with SOTA models} demonstrate the superior performance of our ETSEF method, which promotes a significant improvement in each evaluation setting when compared to the powerful baselines and state-of-the-art methods. With the support of various powerful pre-trained base models, the ETSEF framework shows promising efficiency in representation learning from limited data samples. 

In this work, we focused on supervised transfer learning and contrastive self-supervised learning due to their ease of implementation, domain-agnostic nature, and proven success in natural image classification tasks \cite{jaiswal2020survey, abou2022transfer}. To minimise model implementation difficulty and computational cost, we utilised ImageNet pre-trained weights, which are readily available from TensorFlow and PyTorch libraries to reduce the pre-training cost. The subsequent double fine-tuning process for each base model took an average of 20-40 minutes using TL and 2-3 hours using SSL on personal desktop equipment. We also conducted feature fusion and selection steps to reduce the dimensionality and complexity of extracted features, ensuring an efficient ensemble learning process. Training higher-level ML classifiers with fused features took only 10-40 seconds, making the final deployment quick and efficient.

Beyond improving domain-specific representation learning and the ensemble learning process, our ETSEF method demonstrated effectiveness in out-of-distribution settings. Experimental results (Fig. \ref{OOD task performance comparison}) show that ETSEF outperforms all three baseline ensemble models when facing unseen medical domains, highlighting the significant enhancement in model generalisation ability achieved by combining various pre-training methods. This improvement is crucial for reducing feature domain-specific designs and enhancing model reusability. The following visual analyses of pre-trained base models, powerful baseline models, and the ETSEF model reveal that combining different network architectures with pre-training methods can offset their individual limitations and improve overall model robustness.    

Given the trend towards developing more complex and deeper domain-specific models, we emphasise the importance of using existing TL and SSL approaches to minimise structural modifications and simplify the deployment process. Our focus is on enhancing the extracted feature quality of base pre-trained models through fusion and selection before further ensemble learning steps. By leveraging standard pre-training paradigms, we aim to build an efficient and robust model capable of handling various medical tasks and domains. Compared to large-scale deep models, our method offers reduced computational costs and time while delivering state-of-the-art performance and robust prediction in real-world scenarios.

\subsection{Advantages of ETSEF}

Based on the observed performance and visual analysis of ETSEF, we have summarised the advantages of our ETSEF method as follows:

\begin{itemize}
\item \textbf{Superior Performance in Medical Imaging Tasks:} The primary advantage of ETSEF is its exceptional performance in challenging medical imaging tasks, which is the main objective of this study. Our experiments went through various medical domains, which typically lack large-scale training data, and demonstrated that our ETSEF model achieves state-of-the-art performance. This advantage is crucial in medical diagnostics, as the main requirement in this area is to provide more accurate predictions and reduce the risk of misdiagnoses. 
\item \textbf{Enhanced Robustness:} The enhanced robustness of ETSEF is another significant advantage, largely due to the utilisation of ensemble learning techniques. The feature fusion and selection steps effectively leverage each base model's strengths, leading to more generalised and reliable predictions. Additionally, the majority voting step integrates predictions from multiple ML classifiers, providing a final prediction with the most agreement, thereby reducing the bias risks associated with using a single classifier.
\item \textbf{Flexible Deployment Environment:} ETSEF offers flexibility in the deployment environment, benefiting from the double-fine-tuning technique. This approach allows users to pre-train base models with medium-sized or limited-scale source domain datasets, reducing the risk of negative transfer. Our method enables users to pre-train base models with unlabeled data using the SSL method, followed by ensemble learning steps to achieve better performance in scenarios with extremely limited labelled source data.
\item \textbf{Modular Structure:} The modular structure of ETSEF enhances its adaptability to different domains and tasks. Each pre-trained base model acts as a feature extractor, providing useful features and representations to support higher-level classifier ensemble learning. This reusability of pre-trained models reduces waste of computational costs and accelerates implementation in new environments, making ETSEF versatile and efficient for various applications.
\end{itemize}

\subsection{Limitations of ETSEF}

Meanwhile, we outline the limitations of ETSEF that require attention and future resolution. Also, we provide suggestions for addressing these issues as directions for future research.

\begin{itemize}
\item \textbf{High computational cost of SSL pre-trained models:} As mentioned earlier, SSL pre-trained base models generally require 3-4 times more training time compared to TL methods using the same source domain datasets. This issue is exacerbated when multiple base models are pre-trained for a single downstream task, wasting computational resources and raising deployment costs. To address this, future work could focus on utilising lightweight model structures to build more efficient base models, thereby reducing training costs. Another approach is to leverage the SSL method's advantage in pre-training with multi-modality data and conduct multi-task training to enhance each base model's reusability and generalisation capability, thus minimising the pre-trained base models required.
\item \textbf{Lack of explainable tools for higher-level ML classifiers:} The integrated structure of ETSEF poses a challenge in explaining the decision-making process of all components. While various XAI techniques like Grad-CAM, SHAP, and LIME can provide visual support for the pre-trained base models, understanding the learning process of higher-level ML classifiers remains difficult. This is because the input feature data undergo extraction, concatenation, and selection through several techniques, losing their sequential and positional information, which complicates restoring them back to the sample image for visual analysis. In this study, we conducted an ablation study to draw a contribution map for each ML classifier and reveal their influence. However, it raises the difficulty of fully understanding the decision process, and there is a lack of XAI tools to provide more intuitive insights. A potential solution is to use linear XAI tools to analyse the importance of features and their effects on the final prediction. Additionally, developing specifically designed XAI tools to support the model's explainability in this scenario could be a good research direction in the future.
\end{itemize}

\section{Methods}\label{sec4}

ETSEF involves six main training steps: supervised representation learning on the large-scale non-medical dataset, supervised and self-supervised representation learning on an intermediate source domain dataset, supervised fine-tuning on the target domain dataset, feature extraction and fusion from pre-trained base models, feature selection and ensemble learning with higher-level ML classifiers, and majority voting of ML classifiers. The baseline models do not combine as many architectures and pre-training methods as ETSEF. Extended data Fig. \ref{Scratch Baseline workflow} and Fig. \ref{Pre-training baseline workflow} outline the workflows of the three baseline models. The two pre-trained baselines share the same ensemble structure, differing only in the pre-training methods and base models used during the second representation learning stage.

In this section, we provide implementation details of the ETSEF method and baseline models, including the experimental environment setup, the hyperparameter configuration for each DL architecture and higher-level ML classifier, and the additional useful works of the technologies we used. Moreover, we want to explain the design choices made in our study and provide the users with a better understanding of building an efficient ensemble model.

\subsection{Experimental Environment Setup}

The implementation environment for model training is based on Python 3.10.12 on a Windows 11 Pro system under a desktop PC equipped with an Intel (R) Core (TM) i5-12600K CPU at 4.8 GHz, an NVIDIA Geforce RTX 4080 GPU, and 64 GB of RAM. All pre-trained ImageNet weights are publicly available and were downloaded from the Keras and Torchvision libraries.

\textbf{Base network choosing and pre-trained models.}  In this study, we utilised multiple DL architectures and pre-training methods to train our base models. Given the significant impact of base network performance on the final model's effectiveness, we aimed to select base models with superior performance and data efficiency. We used ImageNet classification performance as a reference for choosing base architectures, as it is a widely accepted benchmark for measuring model performance. Initially, we considered seven prominent neural network architectures: ResNet, Inception, VGG, InceptionResNet, MobileNet, NasNet, and EfficientNet. Additionally, we evaluated popular transformer-based models such as Vision Transformer and Swin Transformer.

Among the CNN architectures, we excluded VGG due to its bad performance and Inception due to its similar structure but lower performance when compared with InceptionResNet. We also excluded the transformer-based models as the preliminary experiments indicated that they struggled with extremely limited data samples, even with pre-training methods. Issues such as overfitting, gradient collapse, and the inability to update layer parameters were observed due to the conflict between these models' large training data requirements and the limited target data available. Eventually, we selected Xception, InceptionResNetv2, MobileNetv3, ResNet101, and EfficientNetB0 as our base models. These medium-sized versions of base architectures, such as ResNet101 and EfficientNetB0, were chosen to balance performance and computational cost.

For initialising the base models, we loaded ImageNet \cite{deng2009imagenet} weights into the backbone architectures. ImageNet was chosen as it is one of the most widely used source domains for pre-training methods, providing a generalised knowledge base to prevent base models from overfitting. Additionally, pre-trained ImageNet weights are easily accessible from various Python libraries and can be prepared within a few minutes. The subsequent representation learning steps were designed to learn domain-specific knowledge from intermediate source domain datasets, selected based on the target datasets' feature distribution and imaging techniques. Both the ETSEF model and the baseline models shared the same pre-trained base models to save computational costs and experimental time. This setup also allowed us to compare ETSEF with the baselines under the same conditions, demonstrating the benefits of the integrated method over using a single pre-training method.

\textbf{Higher-level ML classifiers choosing.} In addition to the pre-trained base models, we trained a group of higher-level ML classifiers using the extracted features to learn correlations from feature distributions and make more robust predictions. For the ML classifiers, we considered popular and widely approved architectures, including Logistic Regression (LR), Adaptive Boosting (AdaBoost), Support Vector Machines, K-nearest Neighbours, XGBoost, Random Forest, and Naive Bayes. These classifiers were chosen for their simple structures and efficient statistical methods, enabling them to learn effectively from input data.

Similar to the selection process for the base models, we excluded Logistic Regression due to its lower performance compared to other methods. We also excluded Adaptive Boosting because both it and XGBoost are boosting-based methods, and XGBoost offers better data efficiency and training speed. Ultimately, we selected five ML classifiers: SVM, KNN, XGBoost, RF, and NB. This selection covers various statistical approaches, including boosting, regression, tree-based methods, and probabilistic models, providing a more generalised measurement and robust prediction capability.

\textbf{Data preprocess and augmentation details.} Considering the varying sizes of image samples due to different medical imaging techniques and collection strategies used to form the training datasets, we applied standard data preprocessing to all data within the source and target domain datasets. We resized all image samples to 224 x 224 pixels to meet the input size requirements of the base models. Additionally, we performed one-hot encoding on all labels of the target dataset samples to facilitate the models' decision-making process and simplify probability prediction. For grey-scale images, we extended their colour channel to three by repeating the original channel, ensuring compatibility with the base models' input requirements.

Data imbalance is a common issue in medical domains, and many of our target domain datasets also exhibit significant class imbalance. To address this, we implemented a series of data augmentation strategies to reduce the imbalance and improve the quality of our training data. Specifically, we used rotation, zoom, flip, and blur techniques. The rotation range was set from -15 to +15 degrees, and the zoom range was from 0.8 to 1.0 of the original image size. Both horizontal and vertical flips were applied, and the kernel size for the blur strategy was set to 9.

However, we avoided using complex or specifically designed augmentation strategies as this was not the focus of our study. Instead, we relied on standard augmentation techniques that have been proven effective in previous research to ensure their reliability. Additionally, we limited the use of data augmentation to the training data by only augmenting classes with fewer samples to the size of the class with most samples to maintain the real-world data scale and preserve the authenticity of learned features, thereby supporting the test of the ETSEF method's efficiency in limited real-world clinical scenarios.

\subsection{Hyperparameter and Classification Head Configuration}

Generally speaking, fine-tuning a base model requires more than minimising the loss function with labelled data; it also necessitates a deep understanding of the model's components. The selection of learning hyperparameters and the configuration of fully connected layers can significantly impact the final model's performance. The following sentences provide details of our hyperparameter choices and classification head design.

\textbf{Base network's configuration.} The pre-training and fine-tuning procedures of pre-trained base models are influenced by several hyperparameters, including batch size, optimiser, weight decay, learning rate, and training epochs. We selected the Adam optimiser for the TL base models and set the initial learning rate to 0.001 to prevent the gradient from dropping too quickly and causing overfitting. We set the batch size to 64 to manage memory usage effectively and the training epochs to 50 to ensure sufficient representation learning. Additionally, we used a weight decay of $10^{-5}$ and monitored the training loss to adjust the learning rate if the lowest loss level did not improve within three epochs. We used a similar setup for the SSL base models to maintain consistency and minimise the influence of changing hyperparameters on performance. The Adam optimiser was also used for the SSL base models, with the learning rate initialised at 0.001. The batch size and training epochs remained the same as for the TL base models, with the only difference being the weight decay monitoring, which started directly at 40 epochs instead of supervising training loss.

The fully connected layers, also known as the classification head in classification tasks, have played a crucial role in influencing a model's prediction performance. A typical classification head includes a flatten layer to convert the learned features from the base models into a one-dimensional array, followed by a pooling layer to downsample and highlight significant features. After pooling, a dense layer with numerous interconnected nodes links the pooling layer to the final output layer, establishing the relationships between learned features. The final output layer is usually a linear layer and works by mapping the model predictions to their corresponding labels.

For TL base models, the classification head during both the pre-training and fine-tuning stages includes a flattened layer and a global average pooling layer to process the learned features, followed by three dense layers to refine and connect these features. We utilised another kind of layer called the dropout layer that drops out nodes by randomly setting them to 0 to mitigate overfitting and enhance generalisation. A dropout layer with a dropout rate of 0.3 is used after the dense layers. The final output layer employs the softmax function to select the label with the highest prediction probability. Conversely, the classification head for SSL base models differs between the pre-training and fine-tuning stages. During the pre-training stage, which is unsupervised, the classification head includes a flattened layer, a max pooling layer, and two dense layers but excludes the final linear layer, with performance evaluated directly by the loss function. In the fine-tuning stage, the classification head retains the same layer configuration but incorporates an output layer with a softmax function to enable supervised fine-tuning.

\textbf{Higher-level ML classifiers' configuration.} Different from the base networks, our ML classifiers are based on different statistical methods and require tailored settings for each classifier's hyperparameter. For the SVM classifier, key hyperparameters included penalty, tolerance, loss function, random state, and max iterations. The penalty and the tolerance determine the decision boundary of the SVM classifier, as the penalty is a regularization term that controls the model's complexity and the degree of tolerance specifies how much misclassification the SVM can tolerate. We tested different penalty functions and various tolerance numbers from 0.1 to 0.0001, and we selected the l2 penalty term and 0.001 as the optimal tolerance, as it provided the best average performance across all test cases. We kept the other hyperparameters, like the loss function and max iterations, at their default settings to reduce the model's variable complexity and maintain performance stability.

The XGBoost classifier uses a booster hyperparameter to decide the statistical method; several options exist, including gradient boost tree (gbtree), gradient boost linear (gblinear), and dart. We selected the tree-based booster based on test case performance for its higher performance behaviour. Additionally, we fine-tuned the tree booster’s hyperparameters, including eta (the learning rate) and max depth (the maximum depth of the tree). We tested eta settings ranging from 0 to 1 and chose 0.9 to optimise performance. For max depth, we set it to 10 to balance model complexity and prevent overfitting.

The NB classifier follows a Gaussian distribution and relies more on the data distribution than the hyperparameters. It has two hyperparameters: priors, which set prior probabilities for the classes, and var smoothing, which adjusts the largest variance portion of all features. For this classifier, we used the default settings, as the test experiment shows that they have a minor influence on the final performance.

The RF classifier is another tree-based classifier, and it involves various hyperparameters such as the number of estimators, criterion, max depth, and min-sample-split that affect the classifier's performance. The estimators, criterion, and max depth are tree-specific parameters. Estimators determine the number of trees in the classifier, max depth defines the maximum depth of each tree, and the criterion measures the quality of a split. Additionally, the Min-Sample-Split parameter specifies the minimum number of samples required to split an internal node. Similar to the XGBoost classifier setup, we set the max depth of the tree classifier to 10 and kept the number of estimators at the default of 100 to control the classifier size and computational cost. We also tested min-sample-split values ranging from 1 to 9 and evaluated criteria such as Gini, entropy, and log-loss functions. Ultimately, we chose a min-sample-split value of 3 and the Gini criterion to maximise the classifier's performance.

The KNN classifier is a classic ML method for classification and regression problems, using proximity to group similar data points. The main hyperparameters of a KNN classifier include the number of neighbours, the weight function, and the algorithm. The number of neighbours K determines how many instances are assigned to the same class as the nearest neighbour, and we set the $k = 3$ to reduce bias between feature points and avoid overfitting. The choice of weight function also influences predictions: the uniform weight function assigns equal weight to all points in each neighbourhood, while the distance function weights points by the inverse of their distance, giving closer points more importance. We selected the distance function for our classifier to prioritise closer feature points and centralise the feature group. We also chose "auto" for the algorithm option to allow the classifier to automatically select the most appropriate algorithm based on the input values.

\subsection{Additional Related Work}

In addition to the methods used in this work, numerous research efforts have employed pre-trained and ensemble learning approaches to address data limitations in the medical field. These approaches can serve as valuable inspiration, broadening our perspective and offering additional insights. In this section, we review previous works related to medical studies to provide a deeper context and inform our model design.

\textbf{Transfer learning.} Despite the scarcity of large-scale labelled medical source domains, TL has been widely applied to medical imaging tasks using large generic datasets as source domain \cite{kora2022transfer}. However, research shows that a lightweight model trained from scratch on medical images performs comparably to a pre-trained model on ImageNet \cite{raghu2019transfusion}, although the pre-trained method offers extra benefits such as speeding up convergence and reducing overfitting \cite{azizi2021big}. At the same time, studies indicate that pre-training with medium-sized labelled medical datasets from the same domain yields better performance compared to training from scratch, demonstrating the value of TL in transferring knowledge from in-distribution domains \cite{alzubaidi2020optimizing, zoetmulder2022domain, juan2024development}. Compared to the supervised TL approaches in which labelled data are expensive and time-consuming to obtain, unsupervised TL approaches that use unsupervised pre-training strategies seem to be a better option to make use of large-scale unlabelled medical datasets \cite{chang2017unsupervised}. Nevertheless, traditional unsupervised TL often fails to achieve state-of-the-art performance, whereas the SSL method as a new branch of the unsupervised TL has garnered increasing attention \cite{huang2023self, theodoris2023transfer}. 

Additionally, multi-source TL has achieved notable success, enhancing the model's representation learning by increasing the modalities of patients and datasets \cite{li2019multisource, yang2020multi}. While the double fine-tuning technique used in this work leverages both large-scale generic datasets and medical domain datasets, extending feature generalisation while maintaining domain-specific knowledge \cite{tan2017distant, niu2021distant}. Our approach also takes advantage of this double fine-tuning method to reduce the pre-trained model's dependency on large-scale medical datasets, and we found that utilising the large-scale generic knowledge base allows the use of small medical datasets as intermediate source domains without causing overfitting.

\textbf{Self-supervised learning.} Previous work has categorised the SSL method into three groups, including contrastive self-supervised learning, predictive self-supervised learning, and generative self-supervised learning \cite{krishnan2022self, xie2022self}, with all three groups have been successfully applied to the medical domain \cite{chen2019self}. The primary distinction among these methods lies in their use of different pretext tasks to guide the pre-training process, which in turn offers various advantages and limitations for downstream tasks \cite{albelwi2022survey}. For instance, a typical generative SSL method called Generative Adversarial Network (GAN) \cite{goodfellow2020generative} employs encoders to learn from original images and generate synthetic images based on the learned features. Another set of encoders then classifies the generated images as real or fake, enabling the GAN model to enhance its feature learning through adversarial competition. Additionally, this approach provides a potential solution to address data scarcity issues in the medical domain by generating large-scale synthetic medical images for training purposes. Meanwhile, the predictive SSL method employs puzzle-based pretext tasks to pre-train the model, enabling it to learn extensive knowledge of positional representations from the training data, making this approach particularly suitable for downstream tasks involving medical localisation and prediction.

Contrastive self-supervised methods offer more advantages in classification tasks than predictive and generative self-supervised learning. These methods pre-train models by comparing and distinguishing input data with pseudo labels, enabling them to learn powerful representations that effectively discriminate the original data \cite{wu2021self}. Contrastive self-supervised methods such as RELIC \cite{mitrovic2020representation}, MoCo \cite{chen2020improved}, SimCLR \cite{chen2020big} and SwAV \cite{tian2020contrastive}, have been applied to medical imaging tasks, achieving significant improvements in label efficiency \cite{liao2022muscle, tiu2022expert}.

Moreover, some works have attempted to design domain-specific pretext tasks to achieve better performance \cite{dong2021self, zhang2024self}, while others focus on multi-task and modality methods \cite{taleb2021multimodal, zhang2023knowledge}. Compared to supervised TL methods, SSL methods using an unsupervised paradigm have shown better potential in multi-task and multi-modality learning, as they do not rely on data labels, which enlarges the availability of the training data and avoids increasing the model's complexity. Currently, there is limited work considering combining SSL with ensemble learning methods, and we are the first work to integrate both TL and SSL pre-trained models as the base models to explore their effectiveness with ensemble learning techniques.

\textbf{Ensemble learning.} Novel and high-performance medical image classification pipelines increasingly utilise ensemble learning strategies. While many ensemble learning approaches combined with deep neural networks have been successful in the medical domain \cite{an2020deep, sreelakshmi2023m}, it remains an open question to what extent and which specific strategies are most beneficial for deep learning-based medical image classification \cite{muller2022analysis}. Meanwhile, traditional ensemble strategies using simple ML classifiers have also proven effective in medical tasks \cite{raza2019improving, namamula2024effective}. However, some studies suggest that ensembling deep neural networks may not be necessary because it significantly increases total computational costs due to the need to train multiple networks \cite{ganaie2022ensemble}. Additionally, further ensembling with pre-trained models has been viewed as a waste of computational resources, adding to the training burden.

Despite these concerns, recent work on ensembling pre-trained models has shown significant performance improvements \cite{el2023enhancing, remzan2024advancing}, highlighting the potential of ensemble learning strategies combined with powerful pre-trained models. Based on these previous works, our study filled this knowledge gap by testing the combination of ensemble learning with recently appeared SSL pre-training methods. We also introduce popular TL pre-training methods to complement the SSL approach's limitation. More importantly, the experiment shows the superior performance and robustness of combining ensemble learning with multiple pre-training methods and architectures.

\section{Conclusion}\label{sec5}

In this work, our objective is to develop an effective deep learning strategy to address the challenge of data scarcity in medical imaging tasks. As a result, we present a novel ensemble framework called `ETSEF' that combines pre-trained base models using TL and SSL methods. The ETSEF strategy has demonstrated state-of-the-art performance and robustness when handling limited training data and challenging medical tasks. Additionally, it is convenient and easy to deploy, and it does not need additional modification of the base model architectures. 

We evaluated the effectiveness of using an ensemble learning strategy with pre-training methods by comparing pre-training baseline ensemble models with scratch-trained baseline ensemble models. The results showed that combining pre-trained base models significantly improved the whole ensemble model's final performance, underscoring the value and necessity of incorporating pre-trained methods to enhance model performance. Our findings also highlighted that TL and SSL-based ensemble models have distinct advantages and limitations towards different data environments, which depend on the training data's colour channel, size, background, imaging techniques, synthetic representations and other factors. Thus, we stress the need for careful consideration of the data environment when combining pre-training methods with ensemble learning. 

Furthermore, our experiments explored multiple ensemble techniques, including feature fusion, feature selection, and weighted voting. The results demonstrated that feature fusion and selection techniques are crucial for maintaining feature quality and significantly improving the final performance of the ensemble model. Weighted voting also improved decision-making accuracy and enhanced the overall reliability of the ensemble model, and we recommend that followers consider these useful techniques when building ensemble models.

Due to the limitations of our experiment scale and computational resources, we did not extensively explore multi-task and modality learning for continuously enhancing pre-trained base models' generalisation, nor did we test more pre-training strategies and powerful architectures to strengthen the base models' feature learning ability. In the future, we plan to extend and test ETSEF in other areas that are bothered by limited training data, and to explore other potential techniques to build powerful and efficient ensemble models.

\backmatter

\bmhead{\large{Data availability}}
All the datasets used in this study are based on publicly available datasets, including source domain datasets and target domain datasets. For the dermatology monkeypox classification task, the source dataset \href{https://www.kaggle.com/datasets/dipuiucse/monkeypoxskinimagedataset}{MSID} is under the CC by 4.0 licence and the target dataset \href{https://www.kaggle.com/datasets/joydippaul/mpox-skin-lesion-dataset-version-20-msld-v20}{MSLDv2} is under the CC by-NC 4.0 licence, and both of them can be downloaded from Kaggle library. For the gastrointestinal endoscopy classification task, the source dataset \href{http://www.multimediaeval.org/mediaeval2018/medico/}{Medico2018} is based on a public medical task, and the corresponding data are available to participants and other multimedia researchers without restriction. The target dataset \href{https://www.kaggle.com/datasets/plhalvorsen/kvasir-v2-a-gastrointestinal-tract-dataset}{Kvasirv2} is under the CC BY-SA 4.0 licence and can be accessed from Kaggle library. For the breast ultrasound classification task, the source dataset \href{https://www.kaggle.com/datasets/anaghachoudhari/pcos-detection-using-ultrasound-images}{POCS} and target dataset \href{https://www.kaggle.com/datasets/sabahesaraki/breast-ultrasound-images-dataset}{BusI} are both available from Kaggle library. For the brain tumour detection task, the source dataset \href{https://www.kaggle.com/datasets/ahmedhamada0/brain-tumor-detection}{Br35H} is shared by its author Ahmed Hamada through Kaggle library, while the target dataset \href{https://github.com/SartajBhuvaji/Brain-Tumor-Classification-Using-Deep-Learning-Algorithms}{BTC} is under the MIT licence and can be accessed from the author's Github website. The out-of-distribution test dataset \href{https://www.kaggle.com/datasets/deathtrooper/glaucoma-dataset-eyepacs-airogs-light-v2}{EyePacs} is a publicly available dataset share within Kaggle library. Moreover, the source domain dataset ImageNet Large Scale Visual Recognition Challenge (ILSVRC) that has been used for each base model's pre-training is publicly available via \href{https://image-net.org/download.php}{ImageNet Website}.

\bmhead{\large{Code availability}} 
To train supervised transfer learning base models, guidelines and tutorials are available in open-source repositories, including the \href{https://keras.io/getting_started/}{Keras} library and \href{https://www.tensorflow.org/}{Tensorflow} library. For building self-supervised learning base models, the \href{https://github.com/google-research/simclr}{SimCLR GitHub Website} provides a good example of implementing contrastive self-supervised pre-training with SimCLR. For more information and implementation samples of GradCAM in PyTorch, refer to the \href{https://github.com/jacobgil/pytorch-grad-cam}{Pytorch-GradCAM Website}. For GradCAM in TensorFlow, visit the \href{https://keras.io/examples/vision/grad_cam/}{Keras-GradCAM Website}. Additionally, information on the SHAP explainer is available in the \href{https://shap-lrjball.readthedocs.io/en/latest/generated/shap.DeepExplainer.html}{Deep SHAP Documentation Website}. Detailed source code and implementation samples of our work can be found on our \href{https://github.com/LaithAlzubaidi/ETSEF/tree/main}{GitHub Website} facilitating better understanding and utilisation of our ETSEF method.


\bibliography{sn-bibliography}

\begin{flushleft}\large{\textbf{Acknowledgements}}\end{flushleft}

We thank all the colleagues who provided us with valuable feedback and suggestions.

\begin{flushleft} \large{\textbf{Author contribution}}\end{flushleft} 

\textbf{Zehui Zhao:} Conceptualisation, Methodology,  Validation, Writing: original draft, \textbf{Laith Alzubaidi:} Conceptualisation, Methodology, Resources, Formal analysis, Writing: review \& editing, Validation, Supervision, \textbf{Jinglan Zhang:}Writing: review \& editing, Formal analysis, Supervision \textbf{Ye Duan:} Formal analysis, Writing: review \& editing, \textbf{Yuantong Gu:} Writing: review \& editing, Validation, Funding acquisition. 

\begin{flushleft}\large{\textbf{Consent for publication}} \end{flushleft}

All authors have given consent for publication.

\begin{flushleft} \large{\textbf{Conflict of interest/Competing interests }}\end{flushleft}

All authors have declared that no conflict of interest exists.

\begin{flushleft}\large{\textbf{Funding}}\end{flushleft}

The authors acknowledge the support received through the following funding schemes of the Australian Government: Australian Research Council (ARC) Industrial Transformation Training Centre (ITTC) for Joint Biomechanics under grant (IC190100020).

\appendix
\begin{appendices}

\section{Extended data}\label{secA1}

\begin{figure}[htbp]
\centerline{\includegraphics[width=0.98\textwidth]{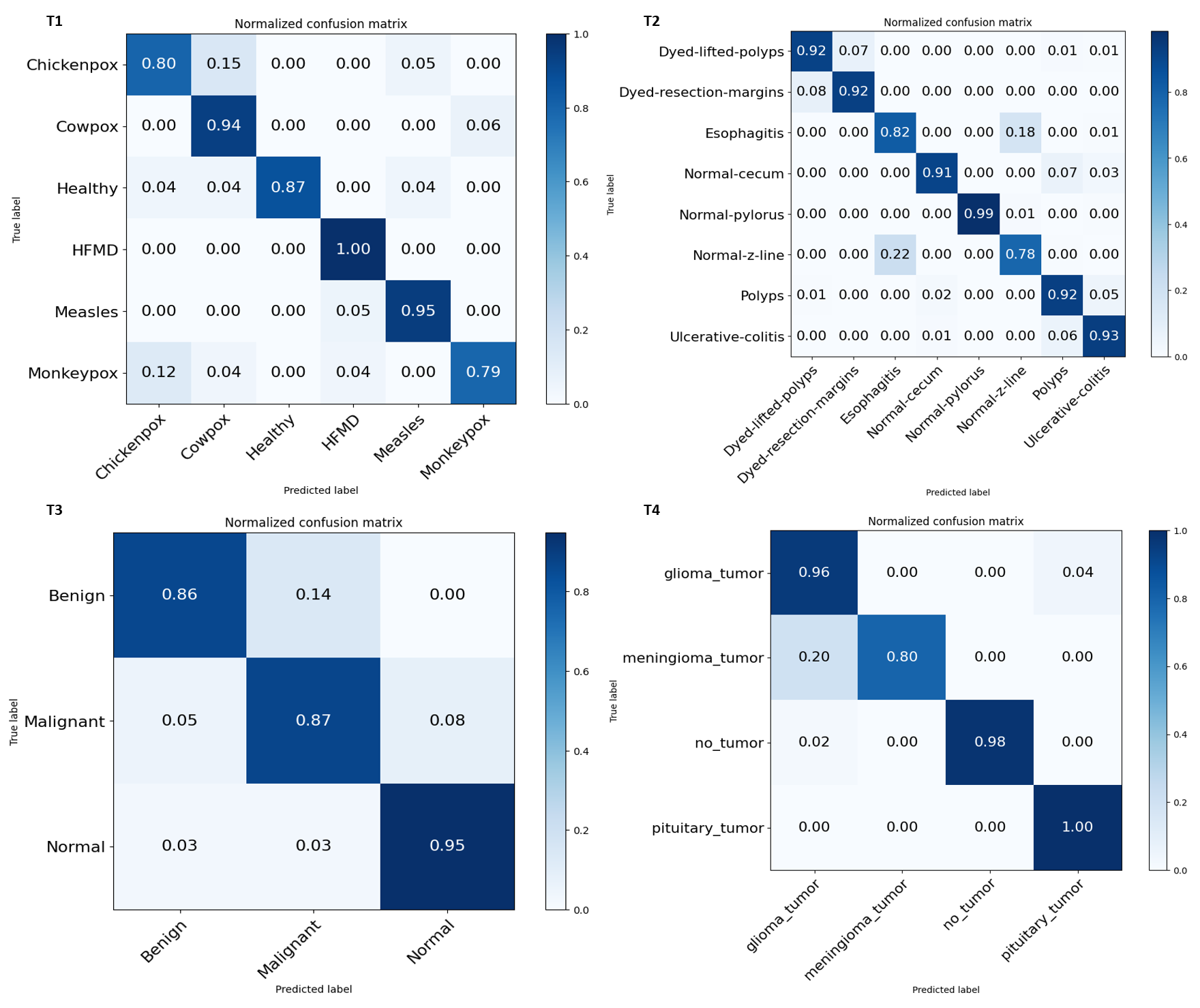}}
\caption{\textbf{Extended Data$|$ Confusion matrix of baseline model without using data augmentation:} This confusion matrix reflects the classification performance of the baseline model on each target task, where all base models were trained directly on the target datasets without pre-training or data augmentation. The matrix reveals significant performance imbalance between classes.}
\label{Confusion matrix baseline}
\end{figure}

\begin{figure}[htbp]
\centerline{\includegraphics[width=0.98\textwidth]{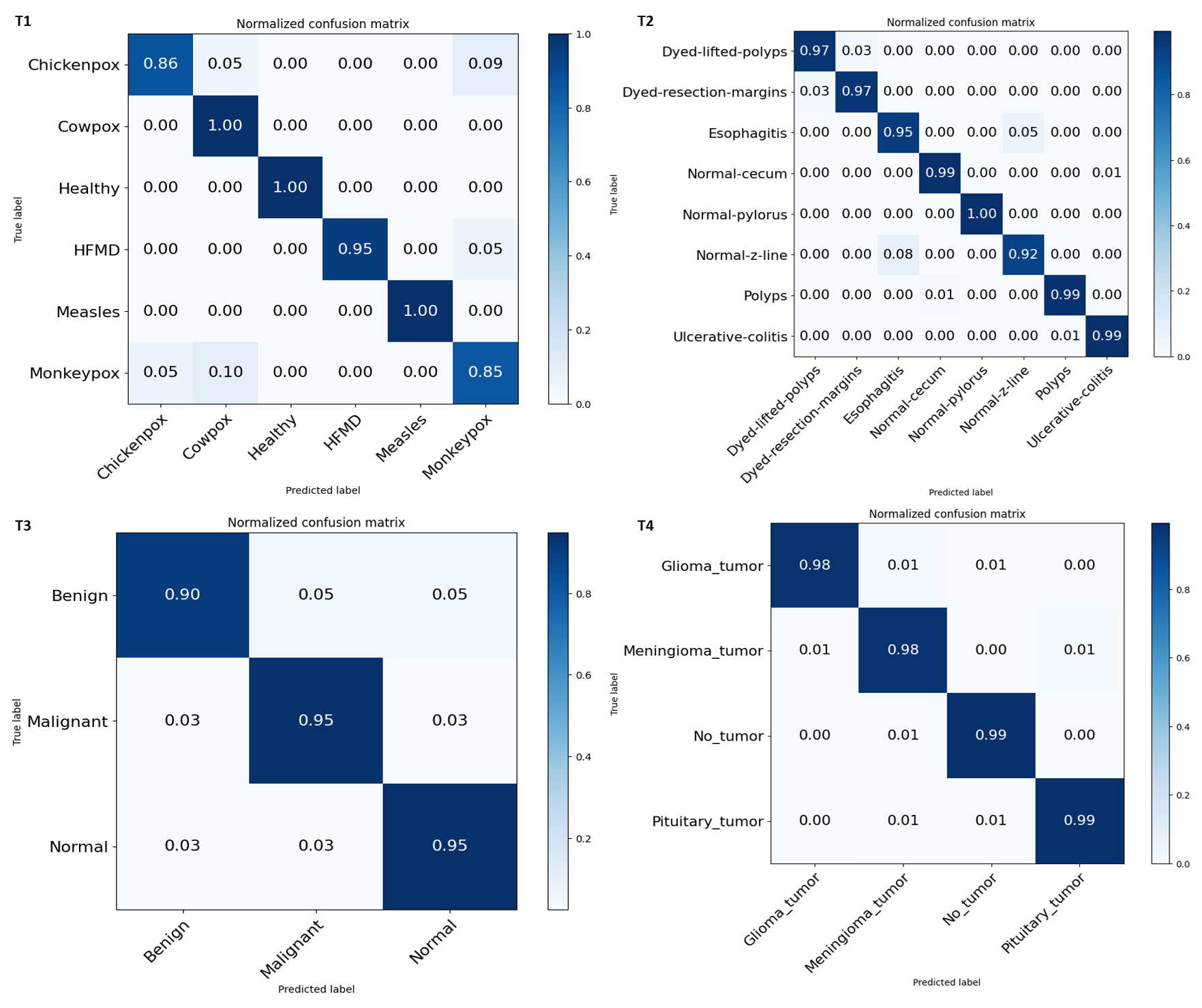}}
\caption{\textbf{Extended Data$|$ Confusion matrix of TL baseline model:} This confusion matrix reflects the classification performance of the TL baseline model, which ensembles pre-trained TL base models. The baseline model demonstrates strong overall performance across each task. }
\label{Confusion matrix baseline TL}
\end{figure}

\begin{figure}[htbp]
\centerline{\includegraphics[width=0.98\textwidth]{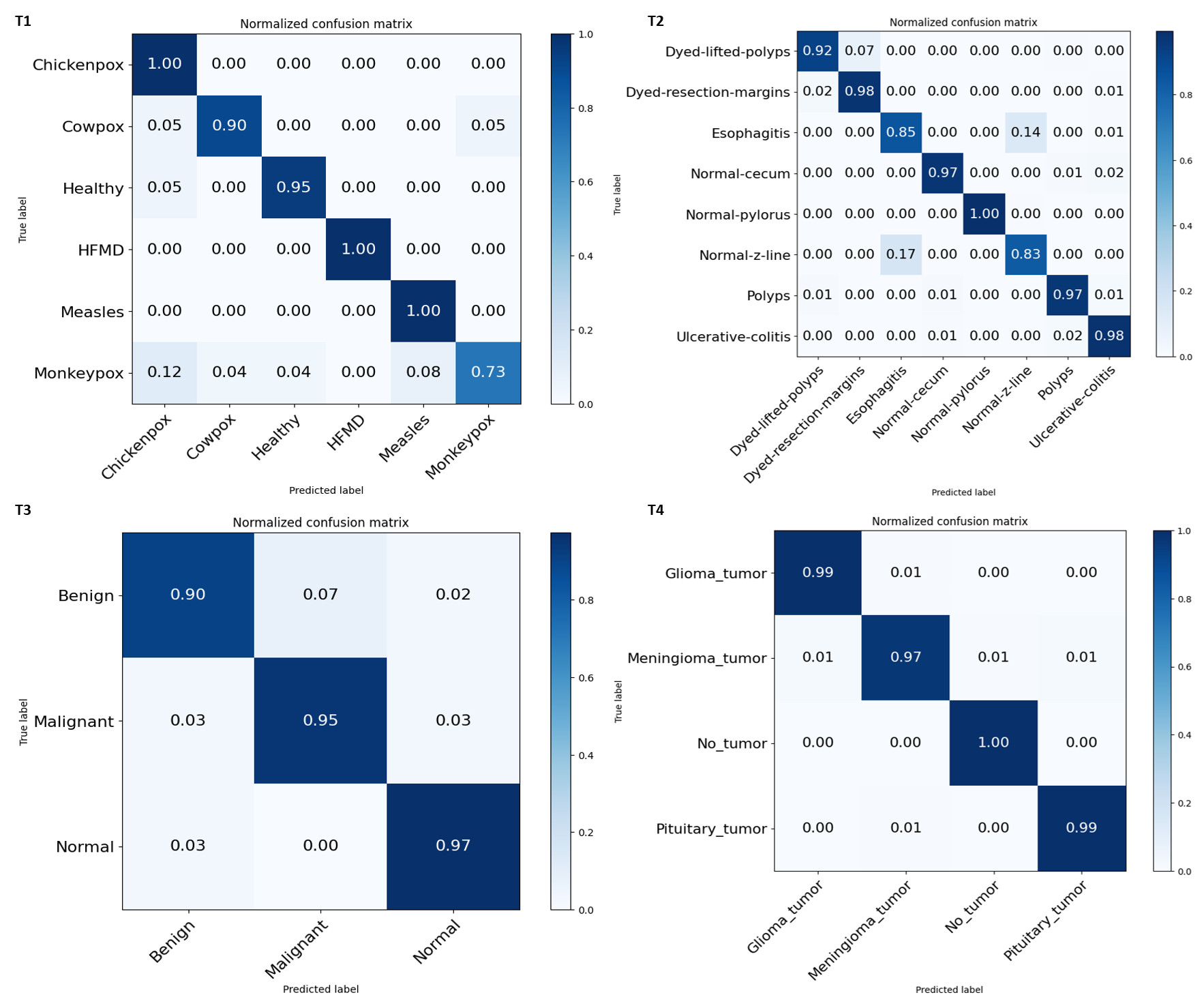}}
\caption{\textbf{Extended Data $|$ Confusion matrix of SSL baseline model:} This confusion matrix reflects the classification performance of the SSL baseline model, which ensembles pre-trained SSL base models. The baseline model demonstrates more balanced performance towards grey-scale image target tasks ($T_3$ and $T_4$) than colourful image target tasks ($T_1$ and $T_2$).}
\label{Confusion matrix baseline SSL}
\end{figure}

\begin{figure}[htbp]
\centerline{\includegraphics[width=0.98\textwidth]{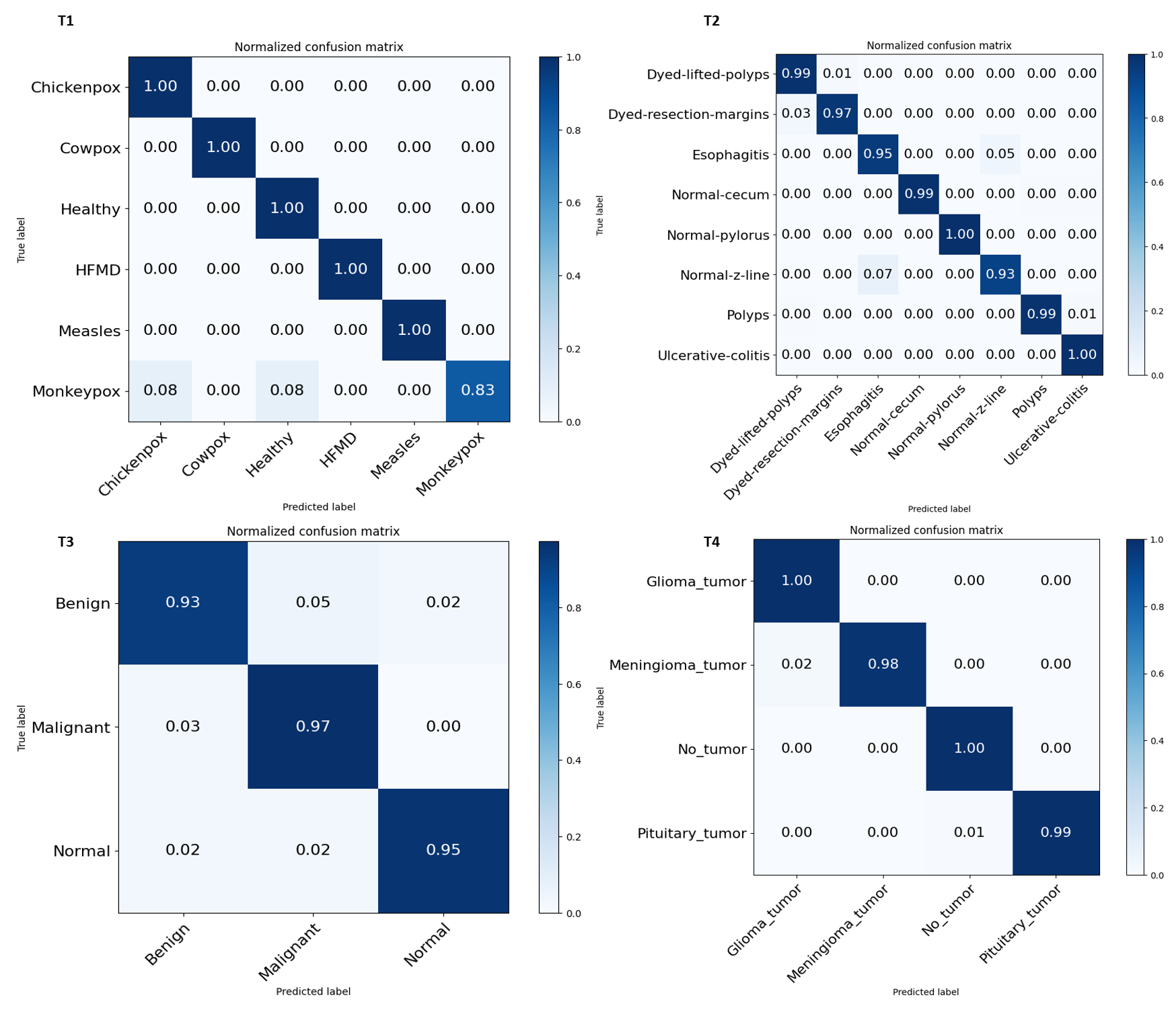}}
\caption{\textbf{Extended Data$|$ Confusion matrix of ETSEF model:} This confusion matrix reflects the classification performance of the ETSEF model, which ensembles pre-trained base models using both TL and SSL methods. The ETSEF model performs best among each target task compared to the three baseline models.}
\label{Confusion matrix}
\end{figure}

\begin{figure}[htbp]
\centerline{\includegraphics[width=0.98\textwidth]{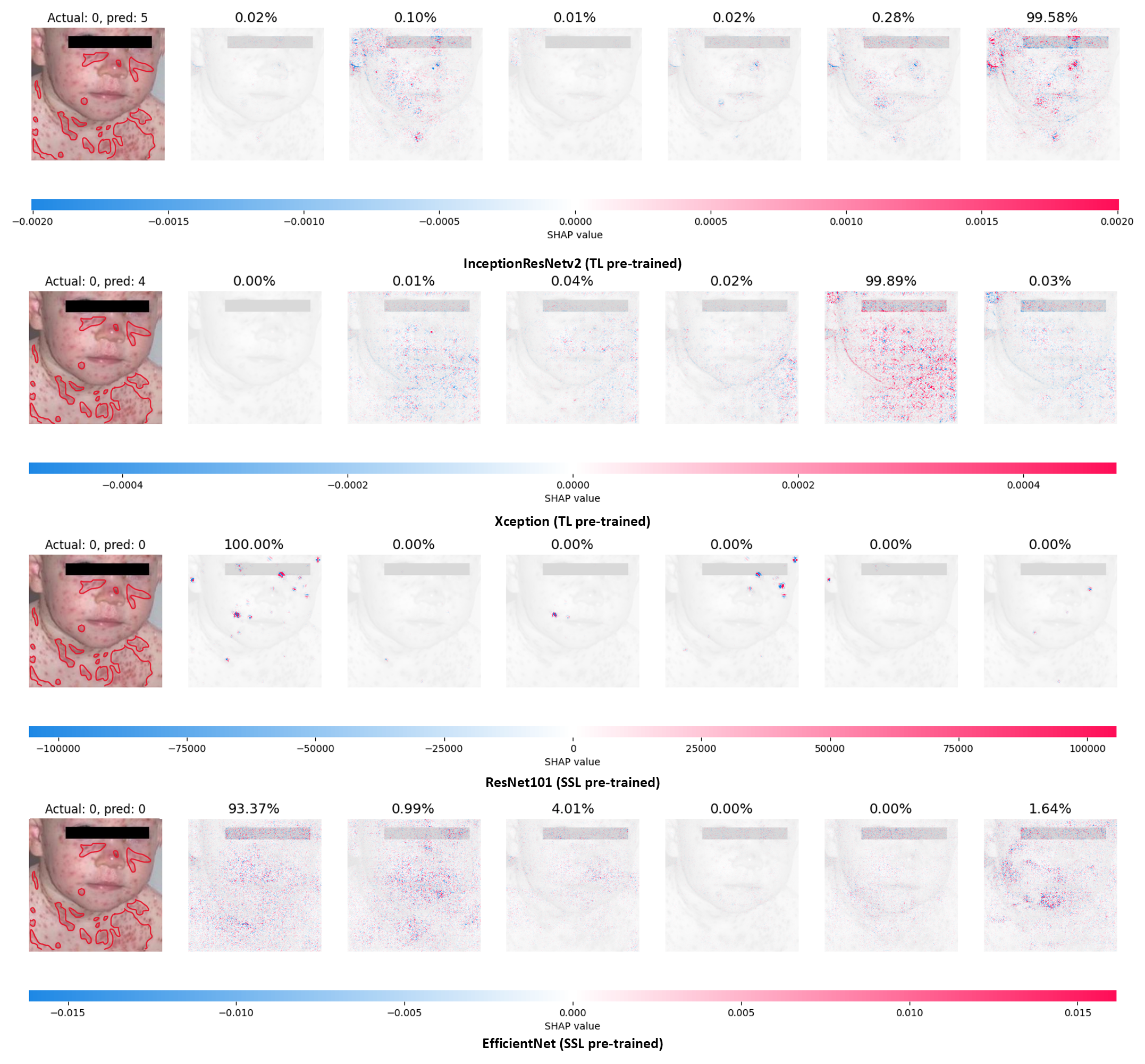 }}
\caption{\textbf{Extended Data$|$ Additional visualisation sample of base models using SHAP maps:} This sample image from $T_1$ demonstrates the differences in attention areas between TL and SSL base models. The TL base models failed to make correct predictions, focusing on unrelated areas, whereas the SSL base models successfully captured part of the symptom area and made correct predictions.}
\label{SHAP map task1}
\end{figure}

\begin{figure}[htbp]
\centerline{\includegraphics[width=0.98\textwidth]{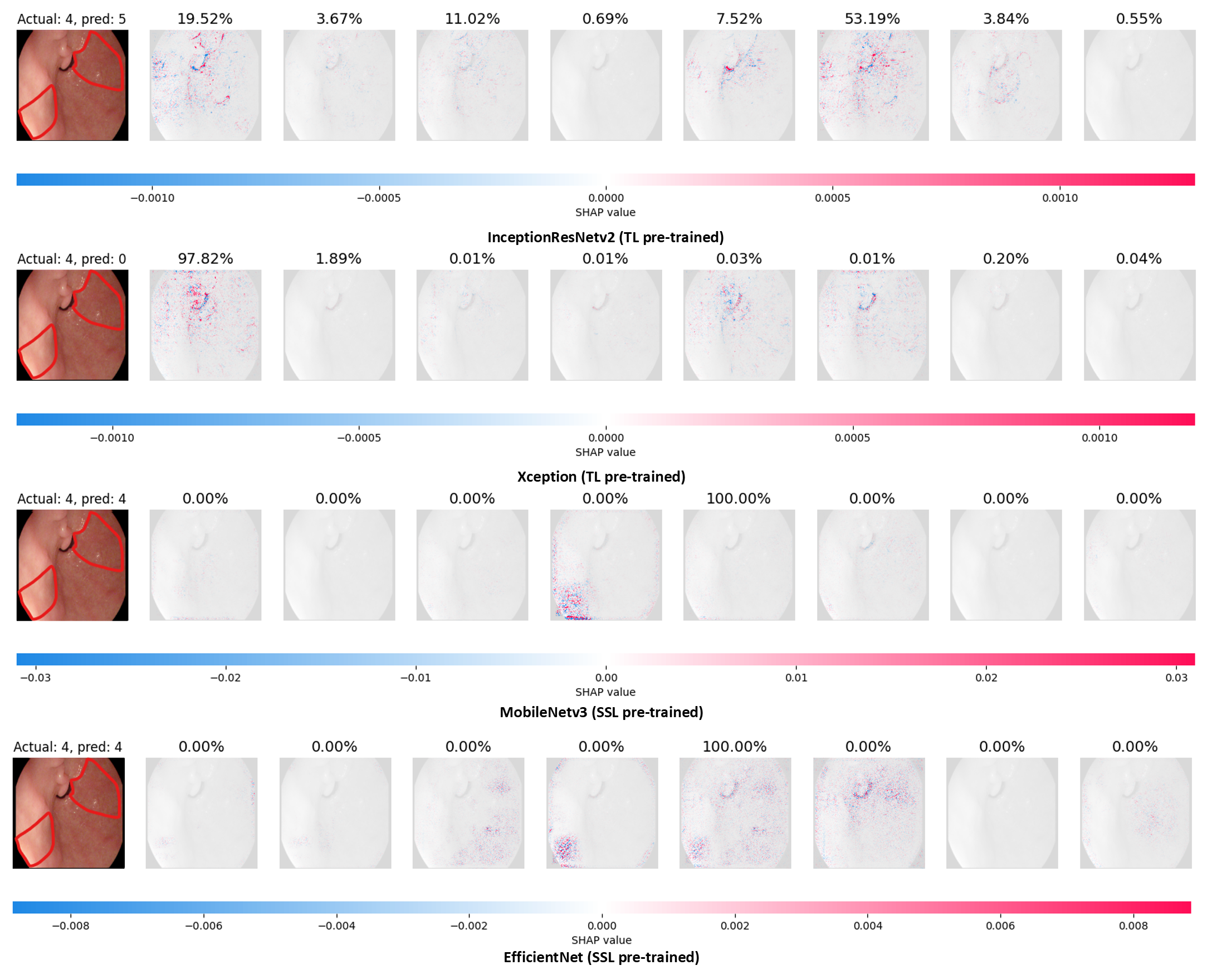}}
\caption{\textbf{Extended Data$|$ Additional visualisation sample of base models using SHAP maps:} This sample image from $T_2$ depicts a normal pylorus image without any disease. The TL base models wrongly focused on unrelated areas and misclassified the sample as a z-line image. In contrast, the SSL model correctly identified the symptoms of a normal pylorus and made accurate predictions based on this understanding.}
\label{SHAP map task2}
\end{figure}

\begin{figure}[htbp]
\centerline{\includegraphics[width=0.98\textwidth]{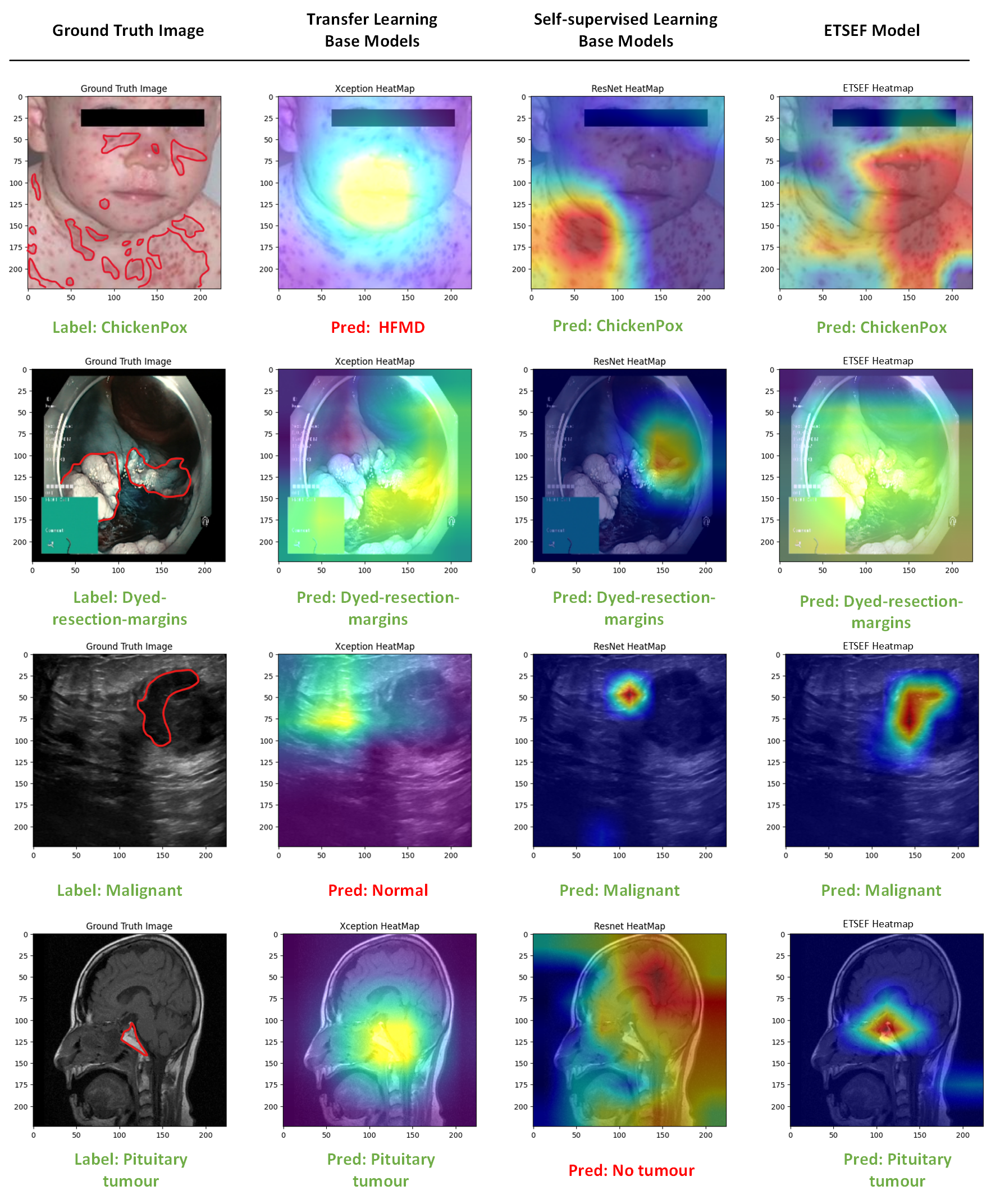 }}
\caption{\textbf{Extended Data$|$ Additional visualisation samples of base models using Grad-CAM maps:} This figure presents additional samples from each task, providing evidence that the ensemble learning technique has enhanced the model's ability to learn and accurately capture relevant representations from the target data. Compared to individual pre-trained base models, the ensemble model extends its attention to cover more symptom areas and provides more precise localisation of disease regions.}
\label{Extend GradCAM heatmap}
\end{figure}

\begin{figure}[htbp]
\centerline{\includegraphics[width=0.98\textwidth]{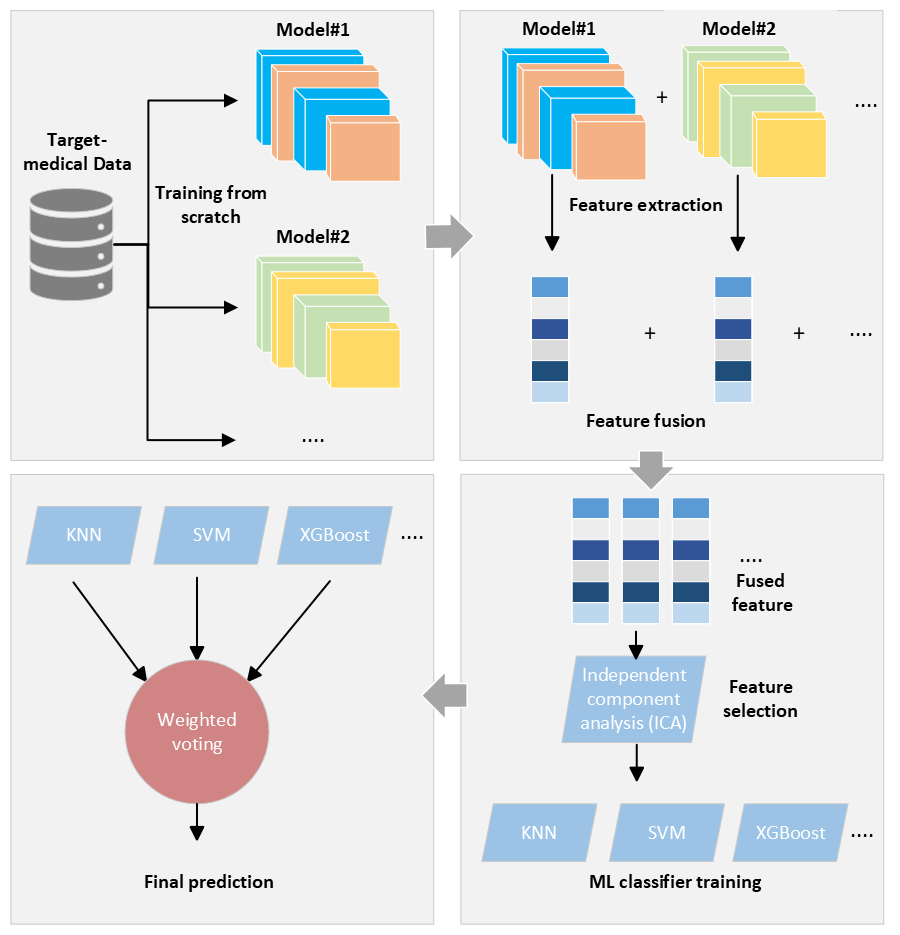}}
\caption{\textbf{Extended Data$|$ Scratch baseline model workflow:} The scratch baseline model employs base models trained directly on the target datasets. Features extracted from each base model are fused and selected through a pipeline. After ensemble learning with ML classifiers, a weighted voting mechanism is applied to ensure the stability and performance of the baseline model.}
\label{Scratch Baseline workflow}
\end{figure}

\begin{figure}[htbp]
\centerline{\includegraphics[width=0.98\textwidth]{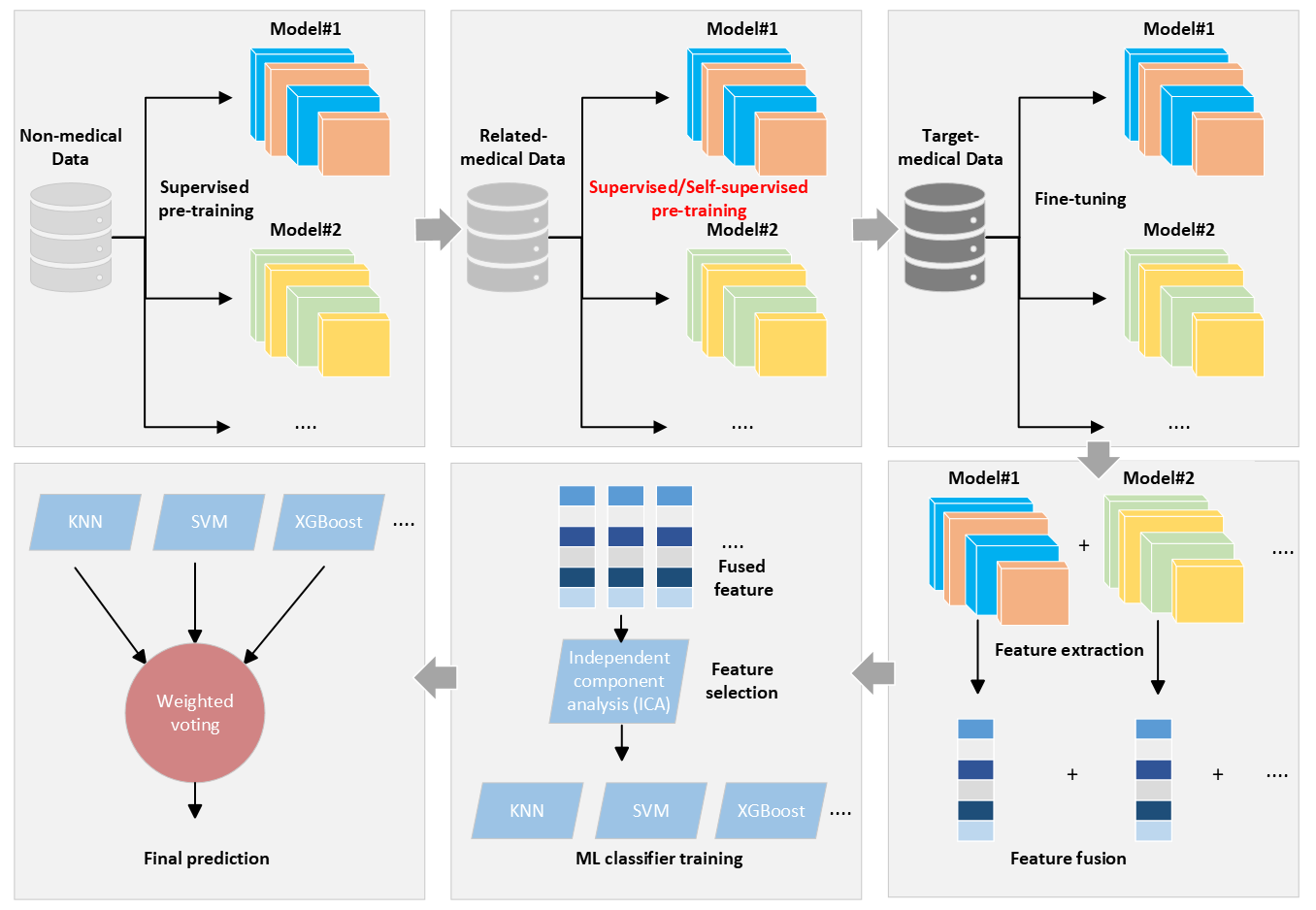}}
\caption{\textbf{Extended Data$|$ Pre-trained baseline model workflow:} The pre-trained baseline model follows a comparable ensemble pipeline to the ETSEF method, differing only in its base models pre-trained with a single pre-training method. These baseline models are constructed for comparison with our ETSEF method and demonstrate the benefits of integrating multiple pre-training methods.}
\label{Pre-training baseline workflow}
\end{figure}




\end{appendices}

\end{document}